\documentclass[authoryear,twocolumn,final,10pt]{elsarticle}

\usepackage{amssymb}

\usepackage{lineno}

\newcommand{\bigO}{\mathcal{O}}
\newcommand{\road}{\mathcal{R}}

\newcommand{\MONTH}{%
	\ifcase\month
	\or January
	\or February
	\or March
	\or April
	\or May
	\or June
	\or July
	\or August
	\or September
	\or October
	\or November
	\or December
	\fi}

\newcommand{\bv}[2]{\mathbf{#1}_\mathrm{#2}} 

\newcommand{\psdmin}{\dot\psi_{\mathrm{min}}}
\newcommand{\psdmax}{\dot\psi_{\mathrm{max}}}

\newcommand{\CSTATE}{\bv{x}{}} 

\newcommand{\states}{\mathcal{X}}

\newcommand{\Xfree}{\mathcal{X}_{\mathrm{free}}}

\newcommand{\REAL}{\mathbb{R}}

\newcommand{\node}{n}
\newcommand{\nodeI}{n_{\mathrm{I}}}

\newcommand{\OPEN}{\textsc{Open}}
\newcommand{\CLOSED}{\textsc{Closed}}	

\newcommand{\sfin}{s_{\mathrm{G}}}

\newcommand{\accmax}{a_{\mathrm{max}}}

\newcommand{\Thor}{T_\mathrm{hor}}

\newcommand{\linmodel}{\mathcal{M}_\mathrm{lin}}

\newcommand{\esm}{\mathcal{M}_\mathrm{ESM}}
\newcommand{\stateESM}{s_{ESM}}
\newcommand{\statesESM}{\bv{s}{ESM}}
\newcommand{\curveradii}{R_\mathrm{c}}

\newcommand{\childnodes}{\bv{n}{}'}
\newcommand{\childcolnodes}{\bv{n}{\mathrm{C}}'}

\newcommand{\feasmap}{f_\mathrm{m}}

\usepackage{footmisc} 										
\usepackage{multicol} 										

\usepackage{subfigure} 										

\usepackage[T1]{fontenc}  								        
\usepackage[applemac]{inputenc}							         
\makeatletter
\let\l@English\l@english
\makeatother
\usepackage[english]{babel} 								        
\usepackage[babel]{csquotes}									
\usepackage{lipsum} 										
\usepackage{url} 											
\usepackage{fancyhdr}										

\usepackage{emptypage}										
\usepackage[english]{varioref} 							                  
\usepackage{microtype}										
\usepackage{eurosym}										
\usepackage{lettrine}                                                                                    
						
\usepackage{amsmath}

\usepackage{mathrsfs}										
\usepackage{amsfonts}										
\usepackage{graphicx}
\usepackage[linesnumbered,ruled,vlined]{algorithm2e}
\usepackage{epstopdf}
\usepackage{upgreek}	
\usepackage{paralist}	
\usepackage{natbib}

\usepackage{mathtools}

\newenvironment{sistema}
{\left\lbrace\begin{array}{@{}l@{}}}								 
	{\end{array}\right.}
\usepackage{amscd}										 
\usepackage{booktabs}										 
\usepackage{graphics} 										 
\usepackage{epsfig} 										 
\usepackage{mathptmx} 										 
\usepackage{times} 											 
\usepackage{rotating}										 
\usepackage{tabularx}										
\usepackage{longtable}										 
\usepackage{pdflscape}										 
\usepackage{listings}										 
\usepackage{xcolor} 										 
\usepackage{siunitx}

\DeclareMathAlphabet{\mathpzc}{OT1}{pzc}{m}{it}
\DeclareMathAlphabet{\mathcal}{OMS}{cmsy}{m}{n}

\usepackage{balance}
\usepackage{tikz}				
\usetikzlibrary{arrows,shapes,backgrounds,calc,positioning,patterns}
\usepackage{multirow}
\usepackage{cases}

\usepackage[moderate]{savetrees}

\usepackage{enumitem} 
\usepackage{textcomp}
\usepackage{pdflscape}
\usepackage{epstopdf}
\usepackage{tikz}
\usepackage{pgfplots} 
\usepackage{graphicx}
\usepackage[export]{adjustbox} 
\usepackage{hyperref}

\usepackage[colorinlistoftodos,prependcaption,textsize=tiny]{todonotes}

\usetikzlibrary{shapes,arrows}
\usetikzlibrary{scopes}
\usepackage{pgf}

\usetikzlibrary{decorations,decorations.markings} 

\journal{Engineering Applications of Artificial Intelligence}

\begin{document}
	
	\begin{frontmatter}
		
		
		
		\title{Search-Based Task and Motion Planning for Hybrid Systems:\\ Agile Autonomous Vehicles }


		\author[add1,add3]{Zlatan Ajanovi\'c}
		\ead{z.ajanovic@tudelft.nl}
		\author[add2]{Enrico Regolin}
		\ead{enrico.regolin@unipv.it}
		\author[add3]{Barys Shyrokau}
		\ead{b.shyrokau@tudelft.nl}
		\author[add4]{Hana \'Cati\'c}
		\ead{hcatic1@etf.unsa.ba}
		\author[add5]{Martin Horn}
		\ead{martin.horn@tugraz.at}
		\author[add2]{Antonella Ferrara}
		\ead{a.ferrara@unipv.it}
		
		\address[add1]{Virtual Vehicle Research GmbH, Inffeldgasse 21/A, 8010 Graz, Austria}
		\address[add2]{University of Pavia, via Ferrata 5, 27100 Pavia,
			Italy}
		\address[add3]{Delft University of Technology, Mekelweg 2, 2628 CD Delft, The Netherlands}
		\address[add4]{University of Sarajevo, Zmaja od Bosne bb, 71000 Sarajevo, Bosnia and Herzegovina}
		\address[add5]{Graz University of Technology, Inffeldgasse 21/B, 8010 Graz, Austria}

		\begin{abstract} 
			To achieve optimal robot behavior in dynamic scenarios we need to consider complex dynamics in a predictive manner. In the vehicle dynamics community, it is well know that to achieve time-optimal driving on low surface, the vehicle should utilize drifting. Hence many authors have devised rules to split circuits and employ drifting on some segments. These rules are suboptimal and do not generalize to arbitrary circuit shapes (e.g., S-like curves). So, the question \lq\lq \textit{When} to go into \textit{which mode} and \textit{how} to drive in it?'' remains unanswered. To choose the suitable mode (discrete decision), the algorithm needs information about the feasibility of the continuous motion in that mode. This makes it a class of Task and Motion Planning (TAMP) problems, which are known to be hard to solve optimally in real-time. In the AI planning community, search methods are commonly used. However, they cannot be directly applied to TAMP problems due to the continuous component. Here, we present a search-based method that effectively solves this problem and efficiently searches in a highly dimensional state space with nonlinear and unstable dynamics. The space of the possible trajectories is explored by sampling different combinations of motion primitives guided by the search. Our approach allows to use multiple locally approximated models to generate motion primitives (e.g., learned models of drifting) and effectively simplify the problem without losing accuracy. The algorithm performance is evaluated in simulated driving on a mixed-track with segments of different curvatures (right and left). Our code is available at \href{https://git.io/JenvB}{https://git.io/JenvB}.
				
		\end{abstract}

		\end{frontmatter}
		

		
		\section{Introduction} 
		\label{sec:intro}

		Similarly to other Artificial Intelligence (AI) applications that can be modeled as Intelligent Agents, Autonomous Vehicle (AV) control is based on a Sensing-Planning-Acting cycle. In this work, we focus on the Planning aspect of the cycle, with the goal to provide feasible Motion Planning (MP) for agile automated driving on a gravel race track. 
		This is not only an exciting problem that attracts a lot of attention in the motorsports, but also a practical benchmark that pushes to the limits our real-time motion planning capabilities.
		
		Existing MP methodologies usually make trade-offs between model complexity and computation time, which is especially challenging in agile automated driving problem. Although simplified vehicle models enable the development of control strategies easier to implement, when the vehicle is driving near the limits of handling (which is the focus of this research), they may fail to represent properly vehicle dynamics. Because of that, many control strategies either generate trajectories that the vehicle can not physically follow or they settle with conservative driving and do not exploit full vehicle possibilities, which is not desirable in racing scenarios.
		This claim is supported by a recent survey on behavior and motion planning for autonomous vehicles by \citet{sharma2021recent}, who claim that future research directions should consider curvature and vehicle orientation as well as tire-road interaction forces. All of these are the major focus of the presented work. 			
		Besides  the obvious use for agile automated driving on a race track, the presented system has relevance for crash avoidance, such that the full vehicle dynamics could be employed to avoid collision with other vehicles \citep{perumal2021insight}.
		
		Due to different dominant effects in vehicle dynamics, we can distinguish different approaches suitable for different road surfaces (e.g., high and low friction coefficients).
		High friction surfaces (high $\mu$) enable better controllability of the vehicle, and the dynamics can be adequately represented with linearized models.  On the other hand, on lower friction surfaces the control action can enter the saturated region where the vehicle dynamics change significantly.

		For driving on high $\mu$ roads, many different methods were presented so far, and some were even tested on real vehicles \citep{valls2018design, betz2019what}.
		Predictive planning of future vehicle trajectories can enable real-time control of driving while avoiding static obstacles \citep{liniger2015optimization}. Recently, \citet{liniger2019noncooperative} extended the approach to racing scenarios with multiple agents (although not real-time). 
		Although these approaches use a nonlinear bicycle model with Pacejka's tire model, the road surface has a high $\mu$, which can be observed as the vehicle is not performing drifting or trail-braking maneuvers.
		Additionally, this MP approach is based on an exhaustive search and works well only for short horizons (due to exponential complexity). Furthermore, it is well suited only for high friction conditions where fast transitions between constant velocity primitives can be achieved. Approaches like this are not well suited for controlling a vehicle in lower friction conditions. Driving in low friction environments requires longer horizons and a more detailed vehicle model, as the control action often enters the saturated region.  Additionally, the solution for minimum-time driving on high $\mu$ surfaces minimizes the curvature of the driving path, so the optimal path is on the edge of the road, which is not robust for a gravel-like road.

		\begin{figure}
			\centering
			\input{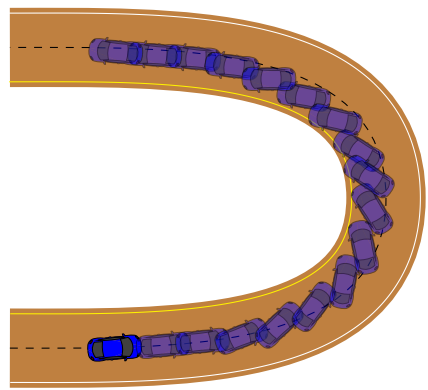}
			\caption{\small Agile automated driving on a slippery road.}
			\label{fig:pd_mopla_prob}
		\end{figure}
  
		Another line of work considers driving on gravel-like roads (low $\mu$), e.g., driving with high side-slip angles like drifting, trail-braking, etc. to improve the robustness of the trajectory.
		Most of the current works in this direction consider two specific scenarios: sustained drift or transient drift.
		One example of a \textit{transient drift} scenario is drift parking, as shown by \citet{kolter2010probabilistic}, where the vehicle approaches the empty parking slot with some velocity and enters temporarily a drift state to rotate and slide laterally in the parking slot.
		On the other hand, in a \textit{sustained drift} scenario, the goal is to maintain steady-state drifting like the well-known phenomena \lq\lq donut drifting'', where the vehicle continuously drifts in circles of small radius.
		Velenis et al. modeled high side-slip angle driving and showed that for certain boundary conditions it can be achieved as a solution to the minimum-time cornering problem \citep{velenis2007modeling, velenis2008optimality}. \citet{tavernini2013minimum} showed that to achieve minimum time cornering with maximum exit velocity in low-friction conditions the vehicle has to go in agile drifting maneuvers.
		Although these works provide deep insight, due to computational complexity, they can not achieve online performance.
		Based on results generated offline, by using the work of \citet{velenis2008optimality}, \citet{you2018real} proposed a method for learning the primitive trail-braking behavior offline, and use the learned model to enable online generation of trail-brake maneuvers. This approach decomposes trail-braking into three stages: entry corner guiding, steady-state sliding, and straight-line exiting.
		A similar approach, based on the decomposition of the problem, was presented by \citet{zhang2018drift}. This approach divides the horizon into three regions, finds a path for each region (using Rapidly-exploring Random Trees (RRT), rule-based sampling, and Proportional Integral Control), and then concatenates them. 
		In these two approaches, decomposition is rather rule-based and not scalable to different circuits.
		Besides simulation work, an impressive demonstration of the scaled vehicle drifting is shown by \citet{williams2017information}, where a Model-Based Reinforcement Learning (MBRL) approach is employed. After extensive trial and error, the RL agent learns the model and, using extensive parallelization, applies planning to find a feasible trajectory and drive on the given track.
		However, as for the aforementioned approaches, the considered scenario is relatively simple with only a single curve. Additionally, this approach demands a lot of experience to learn the model and a lot of computational resources to find a feasible plan.
		An even more impressive demonstration of drifting is presented by \citet{goh2019towards}, where a full-scale DMC DeLorean vehicle is able to drift. However, \citet{goh2019towards} present only a controller that requires a predefined path and not a motion planner. They consider only drifting with a few steady-state drifting options (e.g., right and left), for a given reference path, without straight driving, effectively avoiding the combinatorial problem. Therefore, this approach would have limitations to generalizing outside of the specific track. We use this controller with our proposed planner in this work to improve close-loop robustness in drifting mode.
		The drift control problem was recently also solved using Reinforcement Learning (RL) \citep{cai2020highspeed}. The authors adopted Soft Actor-Critic (SAC), the state-of-the-art model-free deep RL algorithm, to train a closed-loop drift controller. Although they show a satisfactory level  of generalization (e.g., on various road structures, tire friction, and vehicle types), as it is an RL approach, no performance guarantees can be provided. Apart from that, it is still only a controller without any decision-making.
		An alternative approach for deciding when to drift is using Finite State Machines as presented by \citet{acosta2019highly}. However, this is also suboptimal and requires extensive engineering, and provides no guarantees on generalization to other scenarios. 
		A more detailed overview of different approaches in the performance-driving domain is presented by \citet{betz2022autonomous}.

		Driving on arbitrary circuit shapes (e.g., including S-like curves) generally requires to be able to generate trajectories for diverse curves, curves with variable curvature radius, and combinations of right and left curves. Optimal driving then consists of not only a single steady-state drifting maneuver but also close-to-straight driving and steady-state drifting in both directions (i.e., right and left). It is obvious that simple rules do not generalize and the question \lq\lq \textit{When} to go into \textit{which mode} and for \textit{how} to drive in it?'' is an open problem. 
		Solving this question can be considered a combinatorial optimization problem (i.e., NP-hard). Besides the combinatorial nature, to decide on the discrete mode (e.g., drifting, close-to-straight driving), the algorithm requires information on the feasibility of that mode (i.e does there exist a collision-free motion trajectory for that mode). This in turn makes it a class of Integrated Task and Motion Planning (TAMP) problems (hybrid systems), that are shown to be hard to solve optimally in a real-time \citep{garrett2021integrated}.

		As deciding on the modes and generating references for full-circuit driving problems can be considered a combinatorial optimization problem, heuristic search methods can be a well-suited approach.
		In this paper, we present a novel A* search-based approach for generating \lq\lq trackable'' vehicle driving trajectories that exploit full vehicle dynamics.
		The presented A* search-based planner is a modified version of the one presented by \citet{ajanovic2018search}, where a rather simple vehicle model was used to generate vehicle trajectories in complex urban driving scenarios. 
		Here, a complex model of vehicle dynamics is used and motion primitives are generated using two different local approximations of the vehicle based on their feasibility (e.g., using the mode-feasibility-map). A bicycle model is used for maneuvers with small side-slip angle values (e.g., entry and exit maneuvers and close-to-straight driving) and an approximation of the full nonlinear vehicle model based on steady-state drifting is used for cornering maneuvers. 
		The space of the possible trajectories is explored in an automated way, by systematically sampling different combinations of motion primitives, guided by a heuristic search.
		By using different locally approximated models for motion primitives generation, our approach can generate trajectories for arbitrary roads and assign appropriate modes for different segments, effectively overcoming the limitations of state-of-the-art approaches.	
		A limited version of this work has been presented at the International Symposium on Dynamics of Vehicles on Roads and Tracks \citep{ajanovic2019search}. This work focuses more on AI principles rather than vehicle dynamics and it is a significantly extended version of the work, with more mature and well-elaborated algorithms and methods, more realistic experimentation and benchmarking to other approaches.

		Some aspects of our work are related to other well-known approaches in the literature.
		Firstly, our approach for approximation of the full nonlinear vehicle model in steady-state cornering maneuvers is related to the closed-loop prediction approach (CL-RRT) of \citet{kuwata2008motion}. They use a closed-loop system model to generate motion primitives when open-loop dynamics are unstable and the exploration by variations of the inputs in the open-loop dynamics becomes inefficient. In our approach, to avoid the same problem, we generate steady-state cornering motion primitives, which are assumed to be executable via closed-loop control, such as ones of \citet{regolin2018multi} or \citet{goh2019towards}. However, the difference is that we collect extensive number of different equilibrium states in a manifold prior to planning and then sample directly from that manifold during the planning.		
		Secondly, it is worth mentioning that besides motion planning for agile automated driving (as shown here), search-based planning was used for automated driving in unstructured environments \citep{montemerlo2008junior_short, adabala2020multi} and urban automated driving \citep{ajanovic2018search}. As well as for other challenging problems like planning footsteps for humanoid robots \citep{ranganeni2020effective}, robot manipulation \citep{mandalika2018lazy}, underwater vehicles \citep{youakim2020multirepresentation} and the aggressive flying of UAVs \citep{liu2018search}. Different from these works, we introduce mode-feasibility-map that enables us to utilize multiple local model approximations and improve planning performance.
		Finally, our mode-feasibility-map resembles the initiation set of options framework \citep{sutton1999mdps} in hierarchical reinforcement learning or preconditions in PDDL \citep{mcdermott1998pddl} and STRIPS \citep{fikes1971strips} planning languages. Different from planning languages, we deal also with continuous dynamics instead of only logic. And different from Hierarchical RL, we employ mode-feasibility-map with planning algorithms.
		
		The paper is structured as follows. Section \ref{sec:problem_models} provides the problem formulation including necessary models as well as performance criteria and formal problem definition.
		In Section \ref{sec:method}, the framework for motion planning and the approach for the generation of motion primitives are presented.
		Experimentation, including details on implementation and simulation study, is presented in Section \ref{sec:exper}.
		And finally, the conclusions and outlook of the work are presented in Section \ref{sec:concl}.

		\section{Problem Formulation} 
		\label{sec:problem_models}

		The goal of this work is to develop a decision-making and control method that achieves a minimum lap time driving on an empty track in low friction conditions, e.g., gravel road. 
		We assume that the vehicle is equipped with a map of the road and a localization system. Therefore, the vehicle has the information about the road ahead, as well as left/right boundaries and exact position and orientation. Moreover, the full vehicle state feedback information is assumed to be available. In particular, besides dynamic states, the low-level controller for state tracking requires measurements and estimations of several quantities, including wheel forces and wheel slips, both longitudinal and lateral \citep{regolin2019sliding}. Finally, the combined longitudinal/lateral tire-road contact force characteristics are assumed to be known and constant.
		The road, on the other hand, is assumed to be empty, flat, with static road-tire characteristic and can have arbitrary shape.
		The typical scenario could be driving in sharp curves (e.g., with radius 15m) and entering high side-slip angle states (drifting) as shown in Figure \ref{fig:pd_mopla_prob}.
		As it can be seen, to achieve minimum lap-time, the vehicle has to go into drifting mode. In a such mode, the vehicle velocity and the longitudinal axis of the vehicle are not aligned, so the vehicle practically slides laterally. 
		 
		Assumptions and requirements for this problem are summarized as follows.
		
		\textbf{Assumptions:}
		\begin{enumerate}[label=A\arabic*]
			\item The AV drives on a slippery road (e.g., gravel road) with known, constant road-tire characteristic.
			\item The AV is equipped with a map of the road and a state estimation (including localization) system.
			\item The AV drives on an empty track.
		\end{enumerate}
		
		\textbf{Requirements:}
		\begin{enumerate}[label=R\arabic*]
			\item The AV should drive safely on the road while aiming for the minimum lap time.
			\item The AV should be capable to drive in arbitrary planar road geometry (e.g., varying curvature radius, mixed right and left curves) without need for adjustments. 
		\end{enumerate}

		To properly define this problem, several aspects have to be defined including vehicle dynamical model, driveable road and performance criteria.
		
		\subsection{Vehicle dynamical model} 
		
		Vehicle trajectories are generated by concatenating smaller segments of trajectories, the so-called \lq\lq motion primitives'' which are generated based on the model for the vehicle planar motion. Therefore, appropriate vehicle models are essential for the feasibility of the final trajectories.
		Modeling the planar motion of the vehicle for agile automated driving is very challenging as there are multiple aspects to be considered, including longitudinal, lateral, and yaw dynamics, tire-road forces, load transfer, etc.
		For motions that do not push the vehicle to the limits of handling, several simplifications can be used to effectively reduce the problem's complexity. For example, for small side-slip angle motions, a linearized model is accurate enough to be used. 
		On the other hand, for motions that exploit full vehicle dynamics, more detailed vehicle models are required. A deep insight into vehicle dynamics modeling is presented by \citet{tavernini2013minimum}. In this work, we present vehicle dynamics only at the level necessary to introduce the Equilibrium State Manifold (ESM) concept, which is used for the generation of motion primitives in Section \ref{sec:method}. 

\begin{figure}
			\centering
			\usetikzlibrary{calc,quotes,angles}
\usetikzlibrary{calc}

\tikzset{
    car_top/.pic={
    
\filldraw [black, fill=#1] plot[smooth, black, tension=.7] 
	coordinates {(2.3248,0) (2.2967,0.4058) (2.0941,0.7925)(1.7847,0.9958) (1.1874,1.0261) (0.9497,1.0108)}
	.. controls (0.9124,1.143) and (0.8648,1.2352) .. (0.7463,1.1888) .. controls (0.7628,1.0703) and (0.7832,1.0089) .. (0.8202,0.952) .. controls (0.6029,0.9902) and (0.1582,1.0079) .. (-0.2536,0.9816) .. controls (-0.3325,1.0605) and (-0.4289,1.0459) .. (-0.5136,0.9787) .. controls (-0.6947,0.9904) and (-0.8232,0.9758) .. (-0.94,0.9524)
	plot[smooth, tension=.7] coordinates {(-0.94,0.9524)(-1.1951,1.0129) (-1.8081,0.9951) (-2.1686,0.8328) (-2.3894,0.4889) (-2.4286,0) };

\fill [fill=#1](-2.4286,0)--(-0.94,0.9524)--(2.3248,0)--(-0.94,-0.9524);

\filldraw  [black, fill=#1] plot[smooth, black, tension=.7] 
	coordinates {(2.3248,0) (2.2967,-0.4058) (2.0941,-0.7925)(1.7847,-0.9958) (1.1874,-1.0261) (0.9497,-1.0108)}
	.. controls (0.9124,-1.143) and (0.8648,-1.2352) .. (0.7463,-1.1888) .. controls (0.7628,-1.0703) and (0.7832,-1.0089) .. (0.8202,-0.952) .. controls (0.6029,-0.9902) and (0.1582,-1.0079) .. (-0.2536,-0.9816) .. controls (-0.3325,-1.0605) and (-0.4289,-1.0459) .. (-0.5136,-0.9787) .. controls (-0.6947,-0.9904) and (-0.8232,-0.9758) .. (-0.94,-0.9524)
	plot[smooth, tension=.7] coordinates {(-0.94,-0.9524)(-1.1951,-1.0129) (-1.8081,-0.9951) (-2.1686,-0.8328) (-2.3894,-0.4889) (-2.4286,0) };

\draw [smooth cycle, black, fill=darkgray](0.52,0.75) .. controls (0.7,0.78) and (1,0.82) .. (1.2,0.85) .. controls (1.7,0.5) and (1.7,-0.5) .. (1.2,-0.85) .. controls (1,-0.82) and (0.7,-0.78) .. (0.52,-0.75) .. controls (0.7,-0.25) and (0.7,0.25) .. (0.52,0.75);
\draw [smooth cycle, black, fill=darkgray](-1.06,0.67) .. controls (-1.33,0.68) and (-1.57,0.67) .. (-1.8,0.68) .. controls (-2.1,0.27) and (-2.1,-0.27) .. (-1.8,-0.68) .. controls (-1.57,-0.67) and (-1.33,-0.68) .. (-1.06,-0.67) .. controls (-1.16,-0.3) and (-1.16,0.3) .. (-1.06,0.67);
\draw [smooth cycle, black, fill=darkgray](0.97,0.87) .. controls (0.32,1) and (-0.84,0.93) .. (-1.47,0.83) .. controls (-1.23,0.8) and (-1.0974,0.7863) .. (-0.88,0.76) .. controls (-0.56,0.75) and (0.02,0.76) .. (0.4,0.78) .. controls (0.63,0.82) and (0.7,0.83) .. (0.97,0.87);
\draw [smooth cycle, black, fill=darkgray](0.97,-0.87) .. controls (0.32,-1) and (-0.84,-0.93) .. (-1.47,-0.83) .. controls (-1.23,-0.8) and (-1.0974,-0.7863) .. (-0.88,-0.76) .. controls (-0.56,-0.75) and (0.02,-0.76) .. (0.4,-0.78) .. controls (0.63,-0.82) and (0.7,-0.83) .. (0.97,-0.87);

\draw [rotate=51, black, fill=red, thin] (-0.75,2.07) ellipse (0.18 and 0.06);
\draw [rotate=-51, black, fill=red, thin] (-0.75,-2.07) ellipse (0.18 and 0.06);
\draw [rotate=-51, black, fill=white, thin] (0.6696,2.0406) ellipse (0.19 and 0.05);
\draw [rotate=51, black, fill=white, thin] (0.6696,-2.0406) ellipse (0.19 and 0.05);
 }}

\begin{tikzpicture}

\pgfmathsetmacro{\y}{4.5}
\pgfmathsetmacro{\x}{3.5}
\pgfmathsetmacro{\Angle}{atan2(\x,\y)}
\pgfmathsetmacro{\AngleR}{30}
\pgfmathsetmacro{\AngleF}{45}
\pgfmathsetmacro{\AngleB}{40}
\pgfmathsetmacro{\AngleDelta}{25}
\pgfmathsetmacro{\COMradius}{0.15}

\coordinate (Origin) at (0,0);
\draw [-latex,thick] (Origin)--++(0,\x+1.5) coordinate (yaxis); 
\draw [-latex,thick] (Origin)--++(\y+1.5,0) coordinate (xaxis); 
\draw [dashed] (Origin)--++(\Angle:6.5cm) coordinate (AngleEnd);
\draw [draw=none] (\y,0) node [below] {} --++(0,\x) coordinate (Fyf);
\draw [draw=none] (0,\x) node [left] {} --++(\y,0);

%

\draw pic["$\psi$", draw=black, text=black, -latex, angle eccentricity=1.25, angle radius=0.8cm]
              {angle=xaxis--Origin--Fyf};

\coordinate (Fyr) at ($ (Origin) + (\Angle:2cm) $);
\node at (Fyr) [rotate=\Angle,draw,thick,rounded corners=1mm,minimum width=1cm, minimum height=0.4cm] {};
\draw [red,-latex,thick] (Fyr)--++(\Angle+\AngleR:1.3cm) coordinate (RedArrowOne);
\draw pic["$\alpha_r$", draw=black, text=red, -latex, angle eccentricity=1.30, angle radius=0.8cm]
              {angle=Fyf--Fyr--RedArrowOne};
\draw [latex-,thick,blue] (Fyr)--++(\Angle-90:0.5cm) node [rotate=\Angle,right] {$F_{y,r}$};           

\draw [thick] (Fyr)--(Fyf); 

\node at (Fyf) [rotate=\Angle+\AngleDelta,draw,thick,rounded corners=1mm,minimum width=1cm, minimum height=0.4cm] {};         
\draw [red,-latex,thick] (Fyf)--++(\Angle+\AngleF:1.3cm) coordinate (RedArrowTwo);
\draw [dashed] (Fyf)--++(\Angle+\AngleDelta:1.3cm) coordinate (DeltaAngleEnd);
\draw pic["$\delta_f$", draw=black, text=black, -latex, angle eccentricity=1.35, angle radius=0.8cm]
           {angle=AngleEnd--Fyf--DeltaAngleEnd};
\draw pic["$\alpha_f$", draw=black, text=red, -latex, angle eccentricity=1.45, angle radius=1cm]
      {angle=DeltaAngleEnd--Fyf--RedArrowTwo};
\draw [latex-,thick,blue] (Fyf)--++(\Angle-90:0.5cm) node [rotate=\Angle,right] {$F_{y,f}$};

\coordinate (COM) at ($ (Origin) + (\Angle:3.7cm) $);
\pic [scale=1.1, rotate=\Angle, opacity=.15] (b) at ($(COM) +(\AngleB:0.2)$){car_top={blue}} ;
\draw [red,-latex,thick] (COM)--++(\Angle+\AngleB:1.3cm) coordinate (RedArrowBeta);
\draw pic["$\beta$", draw=black, text=red, -latex, angle eccentricity=1.25, angle radius=1cm]
      {angle=AngleEnd--COM--RedArrowBeta};
\begin{scope}[rotate=\Angle]
\fill [radius=\COMradius] (COM) -- ++(\COMradius,0) arc [start angle=0,end angle=90] -- ++(0,-2*\COMradius) arc [start angle=270, end angle=180];
\draw [thick,radius=\COMradius] (COM) circle;
\end{scope}
\draw [-latex,thick,green] (COM)--++(\Angle+90:0.5cm) node [left,rotate=\Angle] {$y$};
\draw [-latex,thick,green] (COM)--++(\Angle:0.8cm) node [below,rotate=\Angle] {$x$};

\draw let \p1 = (COM) in node  (xcom) at (\x1,0) [below] {$x$};
\draw [dotted,thick] (xcom)--(COM);
\draw let \p1 = (COM) in node (ycom) at (0,\y1) [left] {$y$};
\draw [dotted,thick] (ycom)--(COM);

\coordinate (LrLabel) at ($ (Fyr) +  (\Angle+90:1cm) $);
\coordinate (COMLabel) at ($ (COM) +  (\Angle+90:1cm) $);
\coordinate (LfLabel) at ($ (Fyf) +  (\Angle+90:1cm) $);
\draw  [{Bar}{latex}-{latex}{Bar}] (LrLabel)--(COMLabel) node (a)[midway,sloped, yshift=6] {$l_r$};
\draw  [{Bar}{latex}-{latex}{Bar}] (COMLabel)--(LfLabel) node (a)[midway,sloped, yshift=6] {$l_f$};

\end{tikzpicture}
			\setlength{\abovecaptionskip}{-0pt}
			\caption{\small Bicycle vehicle model.}
			\label{fig:bicycle_model}
		\end{figure}
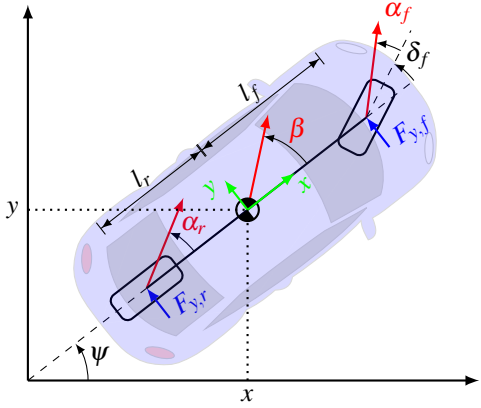

		To describe appropriately vehicle planar motion, the dynamic vehicle model is comprised of six state variables
		$[x, y, \psi, v, \beta, \dot\psi]^{T}$, where $x$, $y$, and $\psi$ represent kinematic states (position coordinates and yaw angle), and $v$, $\beta$, and $\dot\psi$ are vehicle velocity, side-slip angle and rate of change of yaw angle respectively (as seen in Figure \ref{fig:bicycle_model}).
		The evolution of $x$ and $y$ is given by the kinematic relations
		\begin{equation}
			\begin{sistema}
				\dot{x} = v \cdot cos(\psi+\beta)\\
				\dot{y} = v \cdot sin(\psi+\beta)
				\label{eq:kinematic_model}
			\end{sistema}
		\end{equation} 
		whereas the evolution of $v$, $\beta$, $\dot\psi$ are governed by the higher order vehicle dynamic model.
		In regular driving situations (i.e., for small values of $\beta$ and $\dot\psi$), the vehicle dynamics identified by \eqref{eq:kinematic_model} is mostly determined by $v$ and $\dot\psi$. Therefore, motion can be planned by means of linearized vehicle models. 
		For cornering maneuvers, especially on slippery surfaces, such a solution is not suitable anymore, due to the effect of $\beta$ in \eqref{eq:kinematic_model} as well as the complexity of the model that accurately describes the evolution of $\beta$ itself, and might require a full nonlinear vehicle model. 

		For the cornering maneuvers, we use the full-vehicle nonlinear model, where longitudinal and lateral tire-road forces $F_{x},F_{y}$ for each (front and rear) axle are obtained from the normal forces $F_z$ and the combined longitudinal-lateral friction model  \citep{pacejka2012}
		\begin{align}
			\label{eq:forces_definition}
			F_{x,i}=F_{z,i} \mu_x(\lambda_i,\alpha_i),\quad F_{y,i}=F_{z,i}\mu_y(\lambda_i,\alpha_i),
		\end{align}
		where $\mu$ is the friction coefficient, $\lambda$ and $\alpha$ the longitudinal and lateral slips respectively. The front longitudinal force $F_{x,f}$ is zero due to the rear-wheel drive (RWD) configuration.
		
		The nonlinear friction functions take the form of the \textit{Magic Formula} (MF) tire friction model, with an isotropic friction model being used for simplicity \citep{pacejka2012}. This requires the computation of the theoretical slip quantities ($\sigma_j$, $j\in\{x,y\}$), which can be obtained from $\alpha$, $\lambda$ as follows
		\begin{equation}\label{eq:2_8}
			\sigma_{x}=\frac{\lambda}{1+\lambda},\quad \sigma_{y}=\frac{\tan \alpha}{1+\lambda},\quad \sigma=\sqrt{\sigma_x^2+\sigma_y^2}.
		\end{equation}   
	
		Then, the  one-directional friction coefficients are given by
		\begin{align}
			\begin{split}
				\mu_i&=\frac{\sigma_i}{\sigma}D  \sin[C_\lambda \arctan \{\sigma B  - E  (\sigma B-\arctan \sigma B )\}] \mathrm{,}
				\label{eq:mu_def}
			\end{split}
		\end{align}
		for $i \in \{x,y\}$, with $B=1.5289$, $C=1.0901$, $D=0.6$, $E=-0.95084$ being the Pacejka parameters corresponding to gravel. 
		
		The vehicle responses 
		can be obtained from the following system of nonlinear equations, which considers the lateral, longitudinal, and rotating balance equilibrium equations around the vehicle center of gravity (COG):
		\begin{equation}
			\begin{sistema}
				\dot{v} = \dot{\psi}v\beta+\frac{F_{x,r}}{m}     \\
				m v (\dot{\beta}+\dot{\psi}) = F_{y,f}+F_{y,r}\\
				J_z \ddot{\psi} = l_fF_{y,f}-l_rF_{y,r}
				\label{eq:nonlinear_model}
			\end{sistema}
		\end{equation} 
		where $J_z$ is the vehicle inertia around the z-axis, $m$ is the vehicle mass, and $l_f$, and $l_r$ are the distances of the vehicle COG from the front and rear axles respectively. 
		In addition, also longitudinal weight transfer is considered.

		A major limitation that stems from using a full nonlinear vehicle model is the so-called \lq\lq curse of dimensionality'' \citep{bellman1962applied}, which causes a computational explosion when using higher dimensional models in numerical algorithms.
		In fact, if we use the full nonlinear vehicle model (6-dimensional) to generate motion primitives, the computational burden increases excessively (as we need exponentially many samples to properly sample it), thus making it a non-viable option for real-time implementation. Therefore, lower dimensional models are preferred regarding computational requirements.
		To overcome the unnecessary increase in computation requirements while maintaining the accuracy of the model, we develop and employ two local model approximations with respective domains of applicability. These are: 
		\begin{itemize}
			\item Equilibrium States Manifold ($\esm$): a convenient approximation of the full nonlinear model, based on the steady-state drifting phenomenon, applicable during cornering (Section \ref{subsec:v_manifold}).  
			\item Semi-linearized bicycle model approximation ($\linmodel$): applicable in straight-driving/mild-turning scenarios (Section \ref{sec:vehmod_semilin}).
		\end{itemize}

		\subsubsection{Equilibrium States Manifold}
		\label{subsec:v_manifold}
		
		As previously mentioned, \lq\lq donut drifting'' is well known in practice (i.e., exploited by drivers) and investigated in the research (also known as so-called steady-state drifting). We utilize this phenomenon and expand the concept by collecting extensive number of (desirably all) feasible steady-states in a manifold that we call Equilibrium States Manifold (ESM).	
		To obtain such a manifold, we perform offline numeric computations on the full nonlinear vehicle model  (and in other case extensive simulations), and collect the feasible steady-state solutions of the vehicle cornering at different curvature radii \citep{velenis2011steady}. These solutions include the vehicle control inputs (steering wheel angle and rear wheels slip), as well as  vehicle states $v$, $\beta$, $\dot\psi$.
		Assuming a RWD drivetrain configuration, and given different sets of values of the constant control inputs (steering wheel angle $\delta$, driving wheels slip $\lambda$), multiple equilibrium points $\stateESM = [v_{ss}, \beta_{ss}, \dot\psi_{ss}]^{T}$ can be computed for a given constant curvature radii $\curveradii$, by considering the uniform circular-motion relation $\dot\psi = \frac{v}{\curveradii}$, 
		and imposing  the steady-state condition \eqref{eq:derivatives_0} in the vehicle model \eqref{eq:nonlinear_model}.
		
		\begin{equation}
			\label{eq:derivatives_0}
			\dot{v}=\dot\beta= \ddot\psi = 0
		\end{equation} 
	
		A race track is composed of different sections, with varying  curvature radii. Therefore, in order to model steady-state drifting with different radius, we need to compute different equilibrium points. 
		For this reason, the sets $\statesESM(R_{\mathrm{c}_i})$, for different curvature radii $i \in \{1,..,r\}$ are computed and then interpolated into a map $v=f(\beta,\dot{\psi})$, which represents the ESM ($\esm$), as follows.  
		\begin{equation}
		\esm = \{ ( v , \beta , \dot\psi ) \in \REAL^3 \mid \dot{v}=\dot\beta= \ddot\psi = 0 \}.
		\label{eq:ESM}
		\end{equation}
		
\begin{figure*}
			\begin{center}
				\includegraphics[trim={3cm 5.7cm 3cm 6.5cm},clip, width=2.\columnwidth]{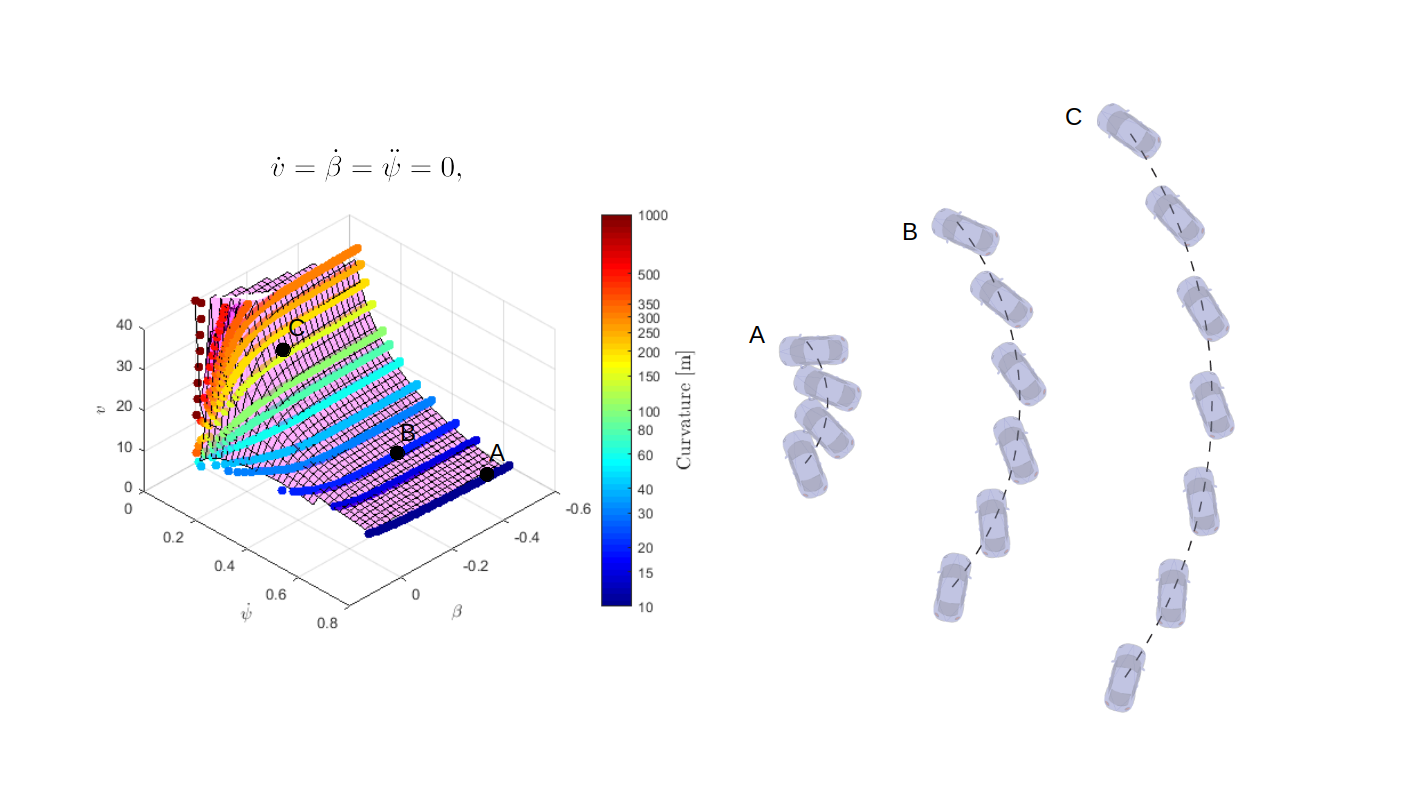}
				\caption{\small On the left, equilibrium points sets $\statesESM$ (and linear interpolation $\esm$) in the $v \times \beta \times \dot\psi$ space for counter-clockwise cornering maneuvers with different curvature radii $\curveradii$. On the right, three segments of motions corresponding to three different states $\stateESM$ on the Equilibrium State Manifold $\esm$ (A, B and C).}
				\label{fig:surface_interp}
				\setlength{\belowcaptionskip}{-10pt}  
				\setlength{\abovecaptionskip}{-10pt}
			\end{center}
		\end{figure*}

\begin{figure}
			\centering
			\subfigure{
				\includegraphics[width=.99\columnwidth]{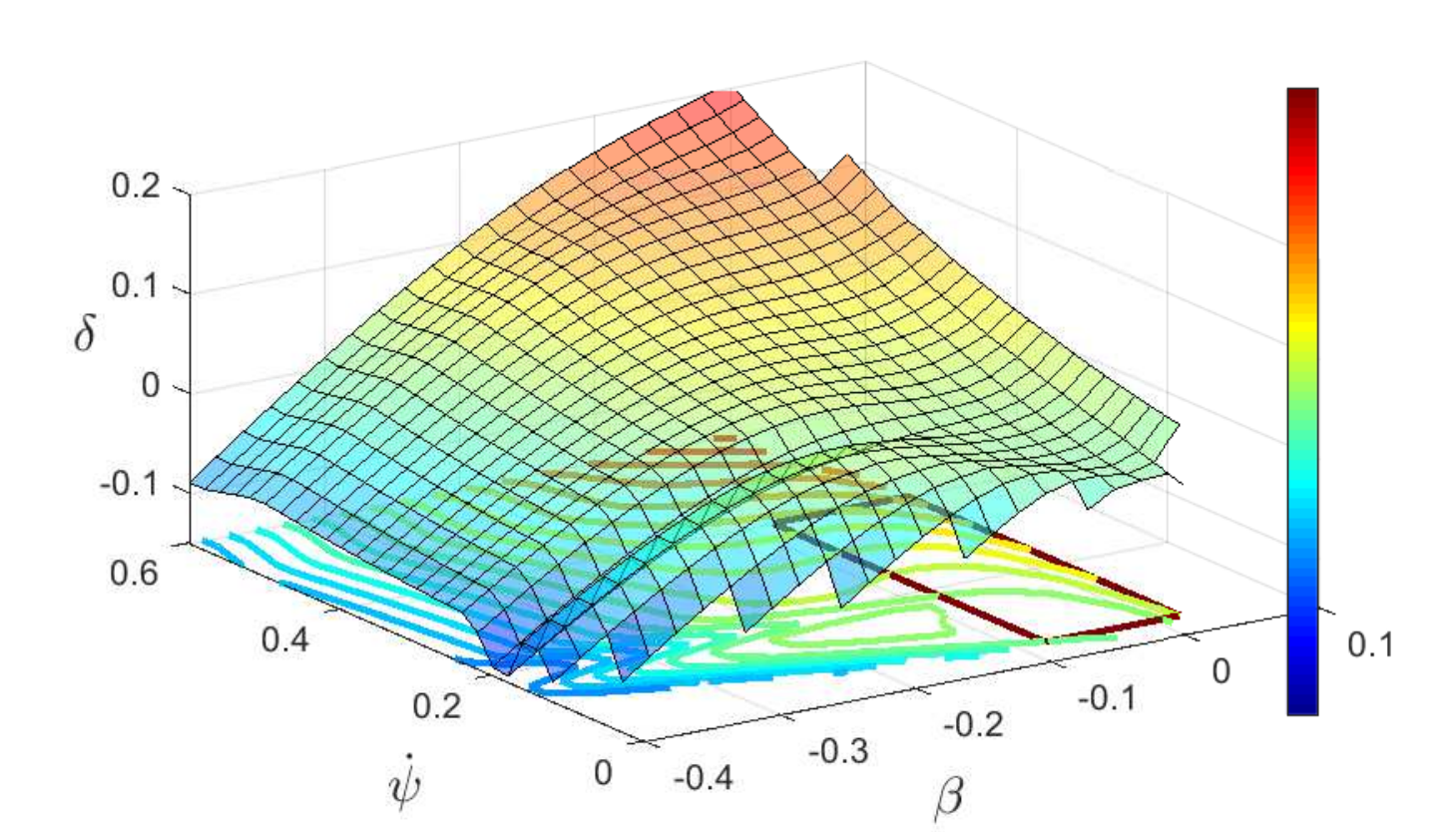}%
			}\hfil
			\subfigure{
				\includegraphics[width=.99\columnwidth]{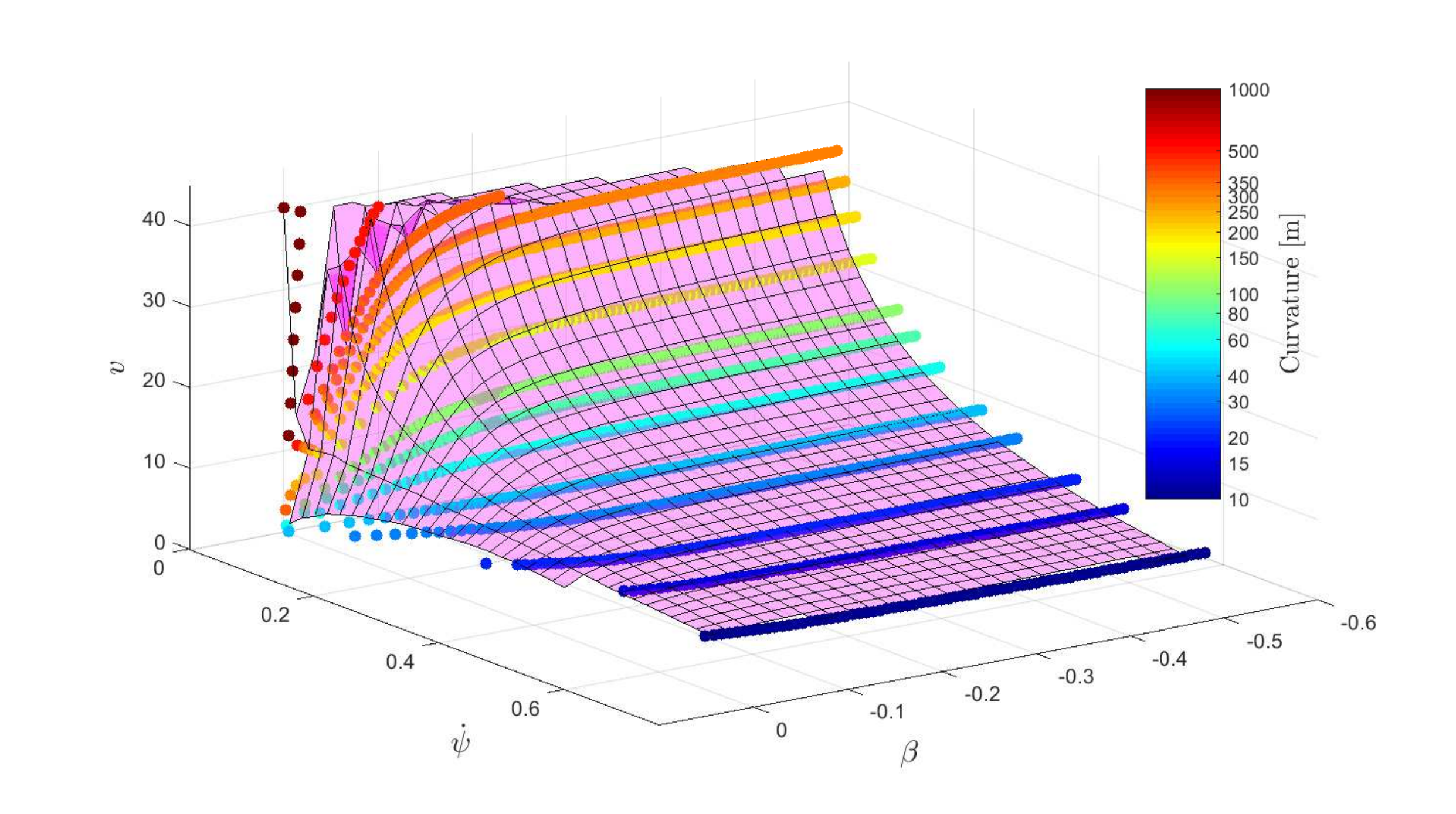}
			}
			\caption{\small ESM for the inputs $\delta$ (top) and $\lambda$ (bottom), counter-clockwise maneuvers. The highlighted portion of the $\beta \times \dot\psi$ plane corresponds to the one in which the bicycle model representation is considered valid. The intervals of values $[\delta_\mathrm{min},\delta_\mathrm{max}]$, $[\lambda_\mathrm{min},\lambda_\mathrm{max}]$ are used for the bicycle-model expansion explained in Section \ref{subsec:node_exp}.}
			\label{fig:surfaces_inputs}%
		\end{figure}

		ESM ($\esm$) is later used to generate steady-state motion primitives by sampling different states $\stateESM$. The same procedure is applied for $\delta$, and $\lambda$.
		In Figure \ref{fig:surface_interp}, these sets $\statesESM(R_{\mathrm{c}_i})$ are visualized in the 3-dimensional state-space for varying $\curveradii$ together with the final interpolated ESM ($\esm$).
		The corresponding surfaces, generated for $\delta$ and $\lambda$ are displayed in Figure \ref{fig:surfaces_inputs}.
		
		Let us assume that the tire-road contact model and the vehicle dynamics model \eqref{eq:forces_definition}-\eqref{eq:nonlinear_model} describe accurately the cornering maneuver dynamics and that a path with curvature radius $\curveradii$ is given, for which at least one reference state $\stateESM(\curveradii)$ exists. Then, if a locally stable feedback controller for the tracking of the state is designed, such path is feasible and can be tracked with appropriate velocity and side-slip angle, given an initial condition close enough to the target state.

		\subsubsection{Semi-linearized bicycle model approximation}
		\label{sec:vehmod_semilin}
		
		When driving conditions are close enough to the origin of the $\beta-\dot\psi$ plane (e.g., close-to-straight driving), a nonlinear bicycle model can be simplified and we can use the semi-linearized bicycle model \citep{genta1997motor}. Therefore, the forces in \eqref{eq:nonlinear_model} can be replaced with their linearized approximations as
		%
		\begin{subequations}
		\label{eq:bicycle_model_forces}
		\begin{align}
		F_{y,f} &= -C_f(\beta+l_f\frac{\dot\psi}{v}-\delta) \\
		F_{y,r} &= -C_r(\beta-l_r\frac{\dot\psi}{v}) \\
		F_{x,r} &= -C_x \lambda .
		\end{align}
		\end{subequations}

		In \eqref{eq:bicycle_model_forces}, the longitudinal and lateral stiffness coefficients $C_x$, $C_f$, $C_r$ are consistent (linearized approximation) with the full characteristics given by \eqref{eq:mu_def} on a given domain.
		This model ($\linmodel$) is valid for range defined by $|\beta|<\beta_\mathrm{lin}$ and $|\dot\psi|<\dot\psi_\mathrm{lin}$ where linearization error is negligible. 

		\subsection{Driveable road} 
		\label{sec:road}
		
		\begin{figure}
			\centering
            \resizebox{1\columnwidth}{!}{%
                \usetikzlibrary{decorations,decorations.markings} 
\pgfkeys{/tikz/.cd,
    street mark distance/.store in=\StreetMarkDistance,
    street mark distance=10pt,
    street mark step/.store in=\StreetMarkStep,
    street mark step=1pt,
}

\pgfdeclaredecoration{street mark}{initial}
{%
\state{initial}[width=\StreetMarkStep,next state=cont] {
    \pgfmoveto{\pgfpoint{\StreetMarkStep}{\StreetMarkDistance}}
    \pgfpathlineto{\pgfpoint{0.3\pgflinewidth}{\StreetMarkDistance}}
    \pgfcoordinate{lastup}{\pgfpoint{1pt}{\StreetMarkDistance}}
    \xdef\marmotarrowstart{0}
  }
  \state{cont}[width=\StreetMarkStep]{
     \pgfmoveto{\pgfpointanchor{lastup}{center}}
     \pgfpathlineto{\pgfpoint{\StreetMarkStep}{\StreetMarkDistance}}
     \pgfcoordinate{lastup}{\pgfpoint{\StreetMarkStep}{\StreetMarkDistance}}
  }
  \state{final}[width=\StreetMarkStep]
  { 
    \pgfmoveto{\pgfpointdecoratedpathlast}
  }
}

\tikzset{
    car_top/.pic={
    
\filldraw [black, fill=#1, very thin] plot[smooth, black, tension=.7, very thin] 
	coordinates {(2.3248,0) (2.2967,0.4058) (2.0941,0.7925)(1.7847,0.9958) (1.1874,1.0261) (0.9497,1.0108)}
	.. controls (0.9124,1.143) and (0.8648,1.2352) .. (0.7463,1.1888) .. controls (0.7628,1.0703) and (0.7832,1.0089) .. (0.8202,0.952) .. controls (0.6029,0.9902) and (0.1582,1.0079) .. (-0.2536,0.9816) .. controls (-0.3325,1.0605) and (-0.4289,1.0459) .. (-0.5136,0.9787) .. controls (-0.6947,0.9904) and (-0.8232,0.9758) .. (-0.94,0.9524)
	plot[smooth, tension=.7,very thin] coordinates {(-0.94,0.9524)(-1.1951,1.0129) (-1.8081,0.9951) (-2.1686,0.8328) (-2.3894,0.4889) (-2.4286,0) };

\fill [fill=#1,very thin](-2.4286,0)--(-0.94,0.9524)--(2.3248,0)--(-0.94,-0.9524);

\filldraw  [black, fill=#1, very thin] plot[smooth, black, tension=.7,very thin] 
	coordinates {(2.3248,0) (2.2967,-0.4058) (2.0941,-0.7925)(1.7847,-0.9958) (1.1874,-1.0261) (0.9497,-1.0108)}
	.. controls (0.9124,-1.143) and (0.8648,-1.2352) .. (0.7463,-1.1888) .. controls (0.7628,-1.0703) and (0.7832,-1.0089) .. (0.8202,-0.952) .. controls (0.6029,-0.9902) and (0.1582,-1.0079) .. (-0.2536,-0.9816) .. controls (-0.3325,-1.0605) and (-0.4289,-1.0459) .. (-0.5136,-0.9787) .. controls (-0.6947,-0.9904) and (-0.8232,-0.9758) .. (-0.94,-0.9524)
	plot[smooth, tension=.7,very thin] coordinates {(-0.94,-0.9524)(-1.1951,-1.0129) (-1.8081,-0.9951) (-2.1686,-0.8328) (-2.3894,-0.4889) (-2.4286,0) };

\draw [smooth cycle, black, fill=darkgray,very thin](0.52,0.75) .. controls (0.7,0.78) and (1,0.82) .. (1.2,0.85) .. controls (1.7,0.5) and (1.7,-0.5) .. (1.2,-0.85) .. controls (1,-0.82) and (0.7,-0.78) .. (0.52,-0.75) .. controls (0.7,-0.25) and (0.7,0.25) .. (0.52,0.75);
\draw [smooth cycle, black, fill=darkgray,very thin](-1.06,0.67) .. controls (-1.33,0.68) and (-1.57,0.67) .. (-1.8,0.68) .. controls (-2.1,0.27) and (-2.1,-0.27) .. (-1.8,-0.68) .. controls (-1.57,-0.67) and (-1.33,-0.68) .. (-1.06,-0.67) .. controls (-1.16,-0.3) and (-1.16,0.3) .. (-1.06,0.67);
\draw [smooth cycle, black, fill=darkgray,very thin](0.97,0.87) .. controls (0.32,1) and (-0.84,0.93) .. (-1.47,0.83) .. controls (-1.23,0.8) and (-1.0974,0.7863) .. (-0.88,0.76) .. controls (-0.56,0.75) and (0.02,0.76) .. (0.4,0.78) .. controls (0.63,0.82) and (0.7,0.83) .. (0.97,0.87);
\draw [smooth cycle, black, fill=darkgray,very thin](0.97,-0.87) .. controls (0.32,-1) and (-0.84,-0.93) .. (-1.47,-0.83) .. controls (-1.23,-0.8) and (-1.0974,-0.7863) .. (-0.88,-0.76) .. controls (-0.56,-0.75) and (0.02,-0.76) .. (0.4,-0.78) .. controls (0.63,-0.82) and (0.7,-0.83) .. (0.97,-0.87);

\draw [rotate=51, black, fill=red, very thin] (-0.75,2.07) ellipse (0.18 and 0.06);
\draw [rotate=-51, black, fill=red, very thin] (-0.75,-2.07) ellipse (0.18 and 0.06);
\draw [rotate=-51, black, fill=white, very thin] (0.6696,2.0406) ellipse (0.19 and 0.05);
\draw [rotate=51, black, fill=white, very thin] (0.6696,-2.0406) ellipse (0.19 and 0.05);
 }}
 
\newcommand{\testpathup}{(-7,0) to[out=0,in=250] (-3.25,1.5) to[out=70,in=180] (-0.5,3)}

\newcommand{\testpathdown}{(0.5,1.5) to[out=0,in=180] (3.25,1.5) to[out=0,in=180] (7,1.5)}

  \begin{tikzpicture}
  [level distance=10mm,
	 valnode/.style={color=red, fill=red, circle, draw,inner sep=1pt,outer sep=0},
	 clnode/.style={color=green, fill=green, circle, draw,inner sep=1.2pt,outer sep=0},
	 colnode/.style={color=red, fill=red, cross out, draw,inner sep=4pt,outer sep=0pt, line width =2},]

\pgfmathsetmacro{\y}{3.5}
\pgfmathsetmacro{\x}{5.5}
\pgfmathsetmacro{\width}{44}
\pgfmathsetmacro{\nsegm}{20}

    \draw[line width=\width, gray] \testpathup; 
    \draw[white,decorate, line width = 0.5,decoration={street mark},street mark distance=\width/2-2] \testpathup; 
    \draw[ draw=white, decorate, line width = 0.5, dashed, decoration={street mark}, street mark distance=(\width-2)/6] \testpathup; 
    \draw[ draw=white, decorate, line width = 0.5, dashed, decoration={street mark}, street mark distance=-(\width-2)/6] \testpathup; 
    \draw[white,decorate, line width = 0.5,decoration={street mark},street mark distance=-\width/2+2] \testpathup; 

   
\foreach \i in {1,...,\nsegm}
{
    \draw[decorate,decoration={markings,
     mark =at position \i*1/(\nsegm+1) with {\draw[black, densely dashed,line width=0.3pt](0,-0.75)
     coordinate(top) -- (0,0.75);}
    }] \testpathup;
 }

  
\coordinate (Origin) at (-7.25,-1);
\draw [-latex,thick] (Origin)--++(0,\y+1.5) coordinate (yaxis); 
\draw [-latex,thick] (Origin)--++(\x+1.5,0) coordinate (xaxis); 

 \node (a) at ($(yaxis)-(0.3,0.3)$){$y$};
 \node (a) at ($(xaxis)-(0.3,0.3)$){$x$};
 


    \draw[line width=\width, gray] \testpathdown; 
    \draw[white,decorate, line width = 0.5,decoration={street mark},street mark distance=\width/2-2] \testpathdown; 
    \draw[ draw=white, decorate, line width = 0.5, dashed, decoration={street mark}, street mark distance=(\width-2)/6] \testpathdown; 
    \draw[ draw=white, decorate, line width = 0.5, dashed, decoration={street mark}, street mark distance=-(\width-2)/6] \testpathdown; 
    \draw[white,decorate, line width = 0.5,decoration={street mark},street mark distance=-\width/2+2] \testpathdown; 

   
\foreach \i in {1,...,\nsegm}
{
    \draw[decorate,decoration={markings,
     mark =at position \i*1/(\nsegm+1) with {\draw[black, densely dashed,line width=0.3pt](0,-0.75)
     coordinate(top) -- (0,0.75);}
    }] \testpathdown;
 }

  
\coordinate (Origin) at (0.25,-1);
\draw [-latex,thick] (Origin)--++(0,\y+1.5) coordinate (yaxis); 
\draw [-latex,thick] ($(Origin)+(0,0)$)--++(\x+1.5,0) coordinate (xaxis); 

 \node (a) at ($(yaxis)-(0.3,0.3)$){$l$};
 \node (a) at ($(xaxis)-(0.3,0.3)$){$s$};

\end{tikzpicture} 
            }
			\caption{\small Curvilinear road representation (Cartesian and Frenet-frame).}
			\label{fig:frenet}
		\end{figure}
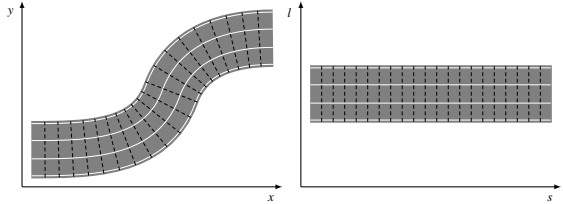
					
		We consider that the vehicle can drive only on the road $\road$. The states outside of the road are considered to be non-driveable states and they are treated as obstacles $(\bigO)$ by the motion planning algorithm. 
		Therefore, the driveability of the trajectory generated based on the vehicle model can be validated based on vehicle coordinates $x$, $y$, and yaw angle $\psi$ only (no need to consider higher dynamical states).						
		As road geometries vary a lot, they can introduce unnecessary complications for motion planning to generate a trajectory that keeps the vehicle on the road. To simplify planning, the driveable road is modeled using a Frenet frame \citep{werling2012optimal}. Instead of using $x$ and $y$ coordinates, in the Frenet frame, one dimension represents the distance traveled along the road $s$, and the other represents the deviation $d$ from the road center-line. By using the Frenet frame, some operations become trivial. For example, to determine whether the vehicle is on the road, it is sufficient to check if the lateral deviation $d$ in the Frenet frame is exceeding half of the road width $w_\mathrm{road}/2$. 
		
		The Frenet frame also ensures that the planning procedure remains the same for each segment of the road. It is important to note that operations in the Frenet frame are used only for trajectory evaluation during planning (e.g., distance traveled, collision checking if the vehicle is on the road, etc.) and grid forming for underlining data structure in planning. On the other hand, the vehicle dynamic model in the Cartesian coordinate system is still used for motion primitive generation. Therefore, we effectively avoid problems of Frenet frame like shown by \citet{li2022autonomous}. Efficient transformations between Frenet and Cartesian frames are necessary as they are used frequently (for every explored node) in each planning step. Figure \ref{fig:frenet} illustrates the procedure of this transformation. Road geometry from the Cartesian coordinate system (left) is represented as a straight road in the Frenet frame (right). 
		Additionally, to ensure all parts of the vehicle are on the road, the vehicle can be represented using multiple circles as shown by \citet{ziegler2010fast}. Ensuring all circles are on the road ensures the vehicle is on the road as well. This is ensured by checking the following condition: 
		\begin{equation}
			\left| d_i \right| \leq  \frac{w_{\mathrm{road}}}{2}-r_i
			\label{eq:road_col}
		\end{equation}
		for each circle $i$ and respective lateral deviation $d_i$ and circle radius $r_i$.
		
		\subsection{Performance criteria}
		\label{sec:obje_mlt}
		
		The goal of the planner is to minimize the time $T$ necessary to drive the full lap.
		In the distance-based formulation, the criteria can be formulated as follows.
		\begin{equation} \label{eq:cost_race}
			T = \int_{0}^{\sfin} \frac{ds}{v(s)\cdot \cos \big(\psi(s) + \beta(s)- \psi_{\mathrm{road}}(s)\big)} .
		\end{equation}
		where term $\psi(s) + \beta(s) - \psi_{\mathrm{road}}(s)$ represents the angle between vehicle velocity and the road tangent.
		So, the whole determinant represents the component of the velocity along the road.
		
		As can be seen, this equation uses the distance as integral bound variables ($0$ and $\sfin$), as it is easier to relate it to the lap start and lap end. The cost function formulated like this can be used to find the global optimal solution. 
		
		However, in MPC, with a fixed time horizon, another formulation can be used to achieve the same effect. Since the goal is to minimize lap time $T$ and the planning time horizon is fixed, equivalent behavior can be achieved by maximizing the distance traveled along the road for a defined time horizon. The criteria can be evaluated simply by considering the first coordinate in the Frenet frame, distance along the path $s$, which is trivial. 
		In the Cartesian frame, this would be equivalently represented as in \eqref{eq:criteria}.
		\begin{equation}
		J_\mathrm{max} = \max_{u(\cdot)} \int_{0}^{t_\mathrm{hor}} v(t)\cdot \cos (\psi(t) + \beta(t) - \psi_{\mathrm{road}}(s))dt.
		\label{eq:criteria}
		\end{equation}

		\subsection{Formal problem definition}
		
		Finally, based on the presented vehicle model and the driveable road we can define the search space that considers kinodynamic constraints imposed by vehicle dynamics:
		
		\begin{equation}
		\begin{split}
			\states & =  \Bigl\{ \CSTATE  \equiv \begin{bmatrix} s, d, \psi, v, \beta, \dot\psi, t \end{bmatrix}^T  \mid (s,d) \in \road, \psi \in \begin{bmatrix} 0, 2\pi \end{bmatrix},\\  
            v& \in \begin{bmatrix} 0, v_\mathrm{max}\end{bmatrix}, \beta  \in \begin{bmatrix}\beta_\mathrm{min},  \beta_\mathrm{max}\end{bmatrix} , \dot\psi \in \begin{bmatrix}\psdmin, \psdmax\end{bmatrix} \Bigl\}.
			\label{eq:PD_searchspace}
		\end{split}
		\end{equation}
		
		As planning is executed in a moving horizon fashion with a fixed time horizon, the goal region is defined as:
		
		\begin{equation}
			\states_G = \{ \CSTATE \mid \CSTATE \in \states, t \geq \Thor\}.
			\label{eq:PD_goal}
		\end{equation}

		Agile automated driving motion planning problem can be formally formulated as follows.
		
		Given:
		
		\begin{itemize} 
			\item the \textbf{search space:} $\states$, ($s \times l \times \psi \times  v \times \beta \times \dot\psi \times t$),
			\item the \textbf{vehicle model:} Equilibrium State Manifold (Section \ref{subsec:v_manifold}), semi-linearized bicycle model (Section \ref{sec:vehmod_semilin}),
			\item \textbf{constraints:}
			\begin{itemize}
				\item \textbf{internal:} acceleration and velocity
				, surface (Section \ref{subsec:v_manifold}), steering and slip (Section \ref{sec:vehmod_semilin}),
				\item \textbf{external:} vehicle is on the road $ |l |\leq \frac{w_\mathrm{road}}{2}$ with is slightly more complex extension \citep{ziegler2010fast} for the full vehicle geometry,
			\end{itemize}
			\item \textbf{objective:} minimum lap-time (Section \ref{sec:obje_mlt}),
			\item \textbf{a query:} initial state $\bv{x}{0}$ and the final state region $\states_G$.
		\end{itemize}
		
		Compute a continuous path $\gamma(\cdot)$ that moves the vehicle from the initial state to the goal region while satisfying all the constraints ($\gamma:[0,1] \mapsto \Xfree$ such that $\tau(0) = \bv{x}{0}$, $\tau(1) \in \states_G $) and minimizing the objective.
		
		\section{Task and Motion Planning Approach} 
		\label{sec:method}

\begin{figure}[t]
        \centering
        \resizebox{\columnwidth}{!}{%
            \usetikzlibrary{positioning,fit,arrows.meta,backgrounds}

\tikzset{
    module/.style={%
        draw, rounded corners,
        minimum width=#1,
        minimum height=7mm,
        },
    module/.default=2cm,
    >=LaTeX
}
\tikzstyle{N} = [circle, minimum size=1mm, draw=black, fill=black,
              inner sep=0pt, outer sep=0pt]
\tikzstyle{line} = [-,>=stealth]

\begin{tikzpicture}[
    show background rectangle]

    \node[module] (I3) {Full model};
    \node[ below=8mm of I3] (andI5) {$\dots$};
    \node[module=0cm, left= 3mm of andI5] (I4) {$f_\mathrm{m}^1$};
    \node[left= 0mm of I4] (andI4) {\&};
    \node[module=0cm, left= 0mm of andI4] (I4) {$\mathcal{M}_1$};
    \node[module=0cm, right= 3mm of andI5] (I6) {$\mathcal{M}_n$  };
    \node[ right= 0mm of I6] (andI6) {\& };
    \node[module=0cm, right= 0mm of andI6] (Ix) {$f_\mathrm{m}^n$};
    \node[ below=3mm of andI5] (mp) {Motion Primitives};
    \node[fit=(I4) (mp)(Ix), draw, inner sep=2mm] (fit1) {};

    \node[ minimum height=10mm, below=16mm of andI5] (tree) {Heuristic function};
    \node[fit=(tree) (tree-|fit1.west) (tree-|fit1.east), draw, inner sep=0mm] (fith) {};
    
    \foreach \i in {4,5,6}
        \draw[->] (I3)--(andI\i.north);

    \node[ minimum height=10mm, right=2cm of {tree-|fit1.east}] (dis) {Discretized Grid};
    \node[right=2.35 of {Ix-|fit1.east}, 
        label={[yshift=5mm] }] 
        (I8) {Tree Search};
        
    \node[N, below=0.1mm of I8](n1){};
    \node[N, below left=1mm of n1](n2){};
    \node[N, below left=1mm of n2](n3){};
    \node[N, below right=1mm of n2](n4){};
    
    \node[N, below right=1mm of n4](n6){};
    \node[N, below left=1mm of n4](n7){};
    \node[N, below right=1mm of n7](n8){};
    \node[N, below left=1mm of n7](n9){};
    \node[N, below right=1mm of n1](n5){};
    \draw[line] (n1) -- (n2);
    \draw[line] (n1) -- (n5);
    \draw[line] (n2) -- (n3);
    \draw[line] (n2) -- (n4);
    \draw[line] (n4) -- (n6);
    \draw[line] (n4) -- (n7);
    \draw[line] (n7) -- (n8);    
    \draw[line] (n7) -- (n9);      
        

    \node[fit={(I8) (I8|-fit1.south) (I8-|dis.west) (I8-|dis.east) (I8|-fit1.north)}, draw, inner xsep=5mm, inner ysep=-\pgflinewidth] (fit8) {};

    \node[fit=(dis) (dis-|fit8.west) (dis-|fit8.east), draw, inner sep=0mm] (fitd) {};


    \draw[<->] (fit1)--(fit8);
    \draw[<->] (fitd)--(fit8);
    \draw[<->] (fith.north east)--(fit8.south west);

    \path (fit1.north east)--node[above=10mm] (arc) {\LARGE{\texttt{SBMP}}} (fit8.north west);

\end{tikzpicture} 
        }
        \caption{\small SBMP Framework.} 
        \label{fig:sbmp_framework}
        \end{figure}
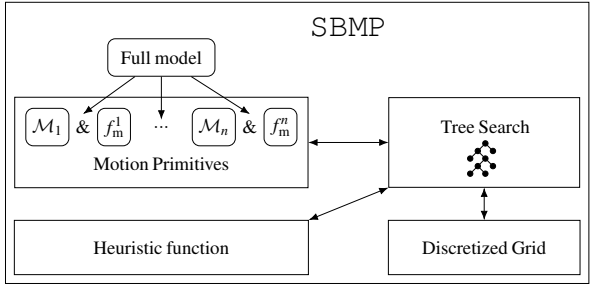

		In this section, we present our search-based task and motion planning framework (SBMP), used for the generation of the driving trajectory. First, we describe some general aspects of the SBMP framework, followed by the clarification of individual components like node expansion and heuristic function, etc. as it can be seen on Figure \ref{fig:sbmp_framework}.

		\subsection{SBMP Framework} 
		\label{sec:framework} 

	\begin{algorithm}[t]
		\DontPrintSemicolon
		\fontsize{8pt}{9pt}\selectfont
		\SetKwData{n}{$n$}
		\SetKwFunction{Search}{Search}
		\SetKwFunction{Expand}{Expand}
		\SetKwFunction{Col}{ColCheck}
		\SetKwFunction{Select}{Select}
		\SetKwFunction{Children}{Children}
		\SetKwFunction{GetTraj}{GetTraj}
		\SetKwInOut{Input}{input}\SetKwInOut{Output}{output}
		\Input{$\nodeI$, $(\road)$, $\linmodel$, $\esm$, $\feasmap$, $h(\node)$}
	\BlankLine
	\Begin{
		$\node \gets \nodeI$\tcp*[r]{initialization}
		$\CLOSED \gets \varnothing$\;
		$\OPEN \gets \node$\;
		\BlankLine		
		\While{$n.k \leq k_\mathrm{hor} $ {\bf and} $ \OPEN \neq \varnothing $ {\bf and}  $ \OPEN.size() \leq N_\mathrm{timeout} $}{
			$n \gets \Select(\OPEN)$\;
			$\OPEN \gets \OPEN \setminus n$\;	
			$\CLOSED \gets \CLOSED \; \cup \; n$\;
			$( \childnodes, \childcolnodes)\gets $  \Expand{$n, \linmodel, \esm, h(\node)$}\; 			
			$\CLOSED \gets \CLOSED \; \cup \;\childcolnodes$\;
			$\OPEN \gets \OPEN \; \cup \; \childnodes$\;			
		}
		\Return{\GetTraj(n)}\tcp*[r]{reconstruct trajectory}
	}
	\caption{\protect\Search: search-based plan for a horizon\label{alg1}}
	\end{algorithm}

	The proposed Task and Motion planning framework is based on the A* search method \citep{hart1968formal}, guided by a heuristic function in an MPC-like replanning scheme.  
	After each time interval $T_\mathrm{rep}$, replanning is triggered from the current vehicle state $\CSTATE$, together with information about the driveable road ahead $\road$. 
	The feasible vehicle trajectories are constructed by concatenating smaller segments of trajectories, the so-called \textit{motion primitives} \citep{frazzoli2002real}.
	The space of the possible trajectories is explored by sampling different combinations of motion primitives in a systematic way, guided by a heuristic search. 
	Motion primitives are generated using two different locally approximated vehicle models. A semi-linearized bicycle model ($\linmodel$) is used for small side-slip angle operations (e.g., curve entry and exit maneuvers and close-to-straight driving) and an approximation of the full nonlinear vehicle model ($\esm$) for steady-state cornering maneuvers.

	The trajectory is constructed by a grid-like search using an A*-like algorithm shown in Algorithm\ref{alg1}. 
	The grid is constructed via equidistant discretization of the state variables $\CSTATE$ in all 7 dimensions. It is important to highlight that the full graph is not constructed in advance, but is built iteratively as the search progresses. In this way, only a small portion of the search space is explored and saved in memory. 
	As we search in the continuous search space $\states$ and expand nodes by sampling multiple motion primitives that generally do not end exactly at grid points. Rounding continuous state to the grid would introduce accumulation of the rounding error. Therefore, an adaptation of the hybrid A* approach \citep{montemerlo2008junior_short} is used for the search. Hybrid A* also uses the grid, but keeps continuous values as well, without rounding it to the grid. When a node is expanded, motion primitives are initiated from the exact continuous state, thus preventing the accumulation of rounding errors. Additionally, keeping only one node in each grid box prunes unnecessary trajectories making it more efficient than purely sampling-based methods.
	
	Each \textit{node} $\node$ contains 20 variables: 6 indexes (representing the grid box) - one for each state in $\CSTATE$ ($\node.x_k$, $\node.y_k$, $\node.\psi_k$, etc.), 6 indexes for the \textit{parent node} (used to reconstruct the solution trajectory at the end of search), six continuous remainders from the discretization of states (used for the initialization of motion primitives) - one for each state in $\CSTATE$ ($\node.x_r$, $\node.y_r$, $\node.\psi_r$, etc.), the exact \textit{cost-to-come} to the node ($\node.g$), and the estimated total cost of traveling from the initial node to the goal region ($\node.f$). The value $\node.f$ is computed as $\node.g + h(n)$, where $h(n)$ is the heuristic function.

	Starting from the \textit{initial node} $\nodeI$ (i.e., representing the initial state), chosen as the first \textit{current node} $\node$. At each iteration, successor nodes are generated in the function $\texttt{Expand}$ by expanding the \textit{current node} $\node$ using motion primitives that are dynamically feasible from that node. The end state from each collision-free motion primitive is represented with one reachable \textit{child node}. All \textit{child nodes} $\childnodes$ are processed and some are added to the $\OPEN$ list. If the \textit{child node} is already in the $\OPEN$ list, and the new \textit{child node} has a lower cost, the parent of that node is updated, otherwise, it is ignored. From the $\OPEN$ list, at every iteration, the node with the lowest cost is chosen to be the next \textit{current node} (in the function $\texttt{Select}$), and the procedure is repeated until the horizon is reached, the whole graph is explored or the computation time limit for planning is reached. At the end of the planning, the node closest to the horizon is used to reconstruct the trajectory.
	
	The planning clearly requires processing time. The compensation for the planning time can be achieved by introducing $T_\mathrm{plan}$, a guaranteed upper bound on planning time. The planning is then initiated from the state $\CSTATE(t+T_\mathrm{plan})$, at which the vehicle would be after the $T_\mathrm{plan}$ time. In this way, the old trajectory is executed while the new one is planned. Thus, the new trajectory is already planned at $t + T_\mathrm{plan}$. This approach has been widely used in MP for automated vehicles \citep{ziegler2014trajectory}.

	\subsection{Node expansion and Motion Primitives} 
	\label{subsec:node_exp}

	\begin{figure}
	\centering
	\input{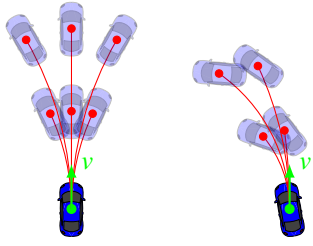} 
	\caption{\small Motion primitives for agile driving using semi-linearized bicycle model $\linmodel$ (left) and ESM $\esm$ (right).} 
	\label{fig:pd_motpr}
	\end{figure}
		
	To build trajectories iteratively, at each iteration of the search, current node $\node$ is expanded, and \textit{child nodes} $\childnodes$ are generated using motion primitives. From each node, $n$, only dynamically reachable and collision-free \textit{child nodes} $\childnodes$ are generated. Each generated child node $\node'$ in  $\childnodes$ represents the end state from one motion primitive trajectory.
	As mentioned before, we employ two types of motion primitives (from two locally approximated models), depending on the \textit{mode-feasibility-map} $\feasmap$. The first type, based on Equilibrium State Manifold $\esm$, is for steady-state drifting during cornering. The second, based on the semi-linearized bicycle model $\linmodel$, is for close-to-straight driving.
	
	Figure \ref{fig:pd_motpr} illustrates motion primitives for two presented models. On the left, motion primitives are generated using a semi-linearized bicycle model $\linmodel$ with 2 variations in the rear wheels slip $\lambda$ and 3 variations in steering wheel angle $\delta$. In total 6 motion primitives are generated. On the left, 4 motion primitives are generated by sampling in ESM $\esm$. In practice, many more motion primitives are generated, up to about 100 successor nodes for each expanded node and around 2000 in total for the whole planning step.
	The complete procedure for the $\texttt{Expand}$ procedure is described in Algorithm \ref{algExp}.
	
	\begin{algorithm}
		\DontPrintSemicolon
		\fontsize{8pt}{9pt}\selectfont
		\SetKwData{n}{$n$}
		\SetKwFunction{Search}{Search}
		\SetKwFunction{Expand}{Expand}
		\SetKwFunction{Col}{ColCheck}
		\SetKwFunction{Sample}{Sample}
		\SetKwFunction{Children}{Children}
		\SetKwFunction{Path}{Path}
		\SetKwInOut{Input}{input}\SetKwInOut{Output}{output}
		\Input{$n$, $\linmodel$, $\esm$, $\feasmap$, $h(\node)$} 
		\BlankLine
		\Begin{
	
		$\childnodes \gets \varnothing$ \;
		$\childcolnodes \gets \varnothing$ \;
		\BlankLine		
		
		
		\tcp*[r]{check applicability based on $\beta-\dot\psi$ map, Figure \ref{fig:feasmap}}
		\BlankLine
		\If{$n \in \feasmap^\mathrm{ESM}$}{
		\BlankLine
		$\childnodes \gets \Sample (n, \esm)$\;		
		}				
		\If{$n \in \feasmap^\mathrm{lin}$}{
		\BlankLine
		$\childnodes \gets \childnodes \; \cup \; \Sample (n, \linmodel)$
		}
				
		$\childcolnodes \gets (\childnodes \mid \childnodes \notin \road) $\tcp*[r]{collision checking}
		
		$\childnodes \gets \childnodes \setminus \childcolnodes$\;

		\BlankLine

		\Return{ $( \childnodes, \childcolnodes)$}\;
	}
	\caption{\protect\Expand: generating child nodes based on motion primitives}
	\label{algExp}
	\end{algorithm}
	
	\subsubsection{Mode-feasibility-map} 
	\label{subsec:feasmap}
	
	Mode-feasibility-map ($\feasmap$) represents domains of feasibility for each of the modes. 
	It is a crucial component that enables solving complex continuous problems as a TAMP and it offers an elegant solution to enable the use of multiple modes or approximated models.	
	Depending on the initial state $\CSTATE$, each of the modes might be feasible or infeasible.
	Mode-feasibility-map resembles the initiation set of options framework \citep{sutton1999mdps} in hierarchical reinforcement learning or preconditions in PDDL \citep{mcdermott1998pddl} and STRIPS \citep{fikes1971strips} planning languages.
	
\begin{figure}
		\centering
		\includegraphics[width=.95\columnwidth]{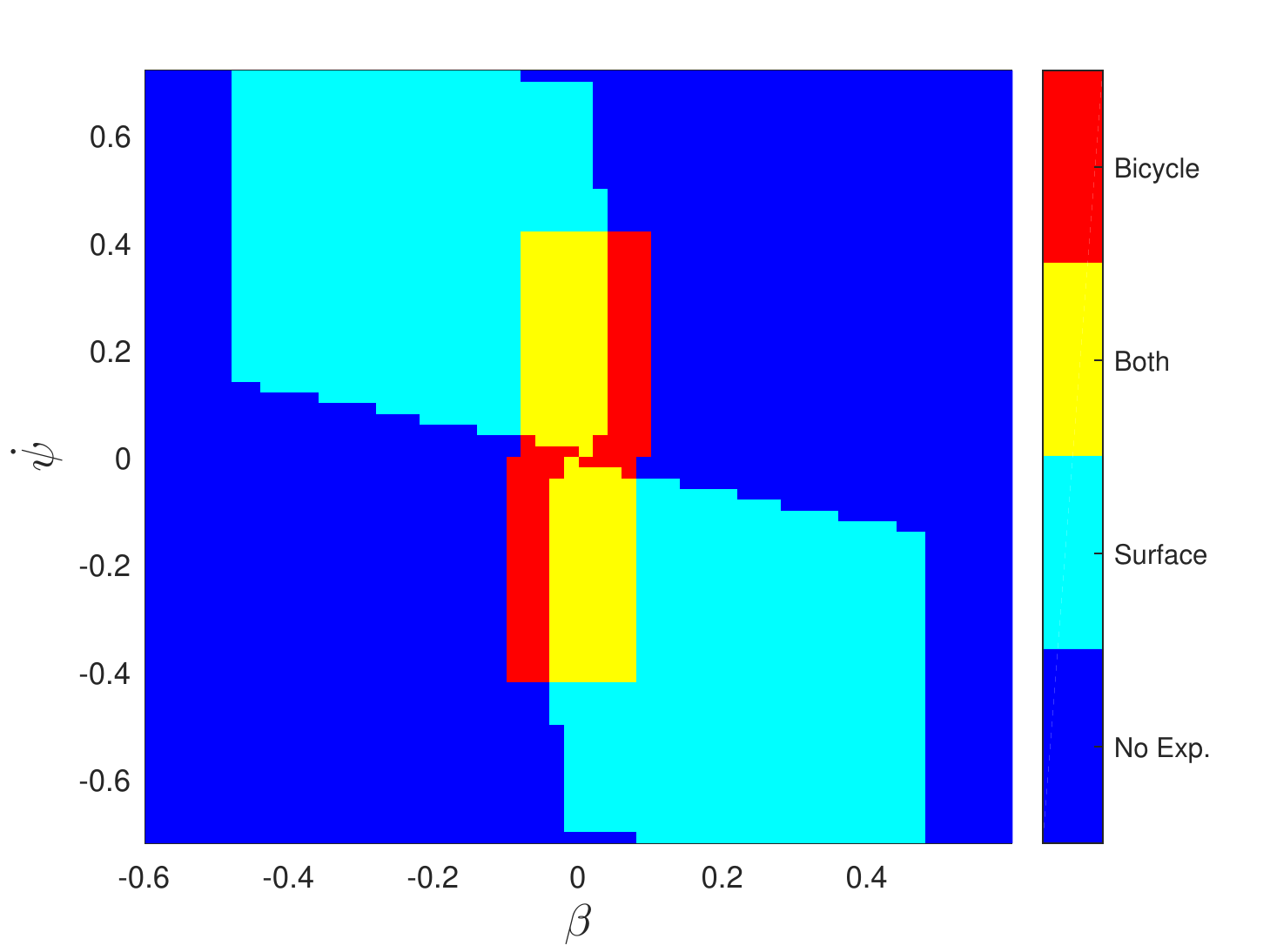}
		\caption{ \small Mode-feasibility-map $\feasmap$, representing feasibly of expansion modes depending on initial states in $\beta \times \dot\psi$.}
		\label{fig:feasmap}%
	\end{figure}

	For our agile driving problem, there are two distinct modes. These are steady-state drifting and close-to-straight driving, as described below. Mode feasibility, in this case, depends only on two states, side-slip angle $\beta$ and yaw rate $\dot\psi$, as in \eqref{eq:feasmap}.	
	\begin{equation}
		\label{eq:feasmap}
		\feasmap(\CSTATE) = \feasmap(\beta , \dot\psi)
	\end{equation} 
	As can be seen in Figure \ref{fig:feasmap}, there are different regions of the $\beta \times \dot\psi$ plane. 	 
	Steady-state drifting is feasible on the regions where Equilibrium State Manifold $\esm$ is defined. As it can be seen Equilibrium State Manifold $\esm$ is symmetric around the origin, as we extended it for drifting in both directions, clockwise and counter-clockwise.	 
	On the other hand, when initial state $\CSTATE$ is close enough to the origin of the $\beta \times \dot\psi$ plane, i.e., for $|\beta|<\beta_\mathrm{lin}$ and $|\dot\psi|<\dot\psi_\mathrm{lin}$, we can employ close-to-straight driving and use $\linmodel$ to generate motion primitives.
	As it can be seen, these two modes complement each other allowing a smooth transition between drifting from one direction to another.	
	Each of the modes is used for generating motion primitives and respective child nodes whenever it is feasible.

	\subsubsection{Steady-state drifting mode}
	
	\begin{figure}
		\centering
		\includegraphics[trim={1cm 0 1cm 0},clip,width=1.\columnwidth]{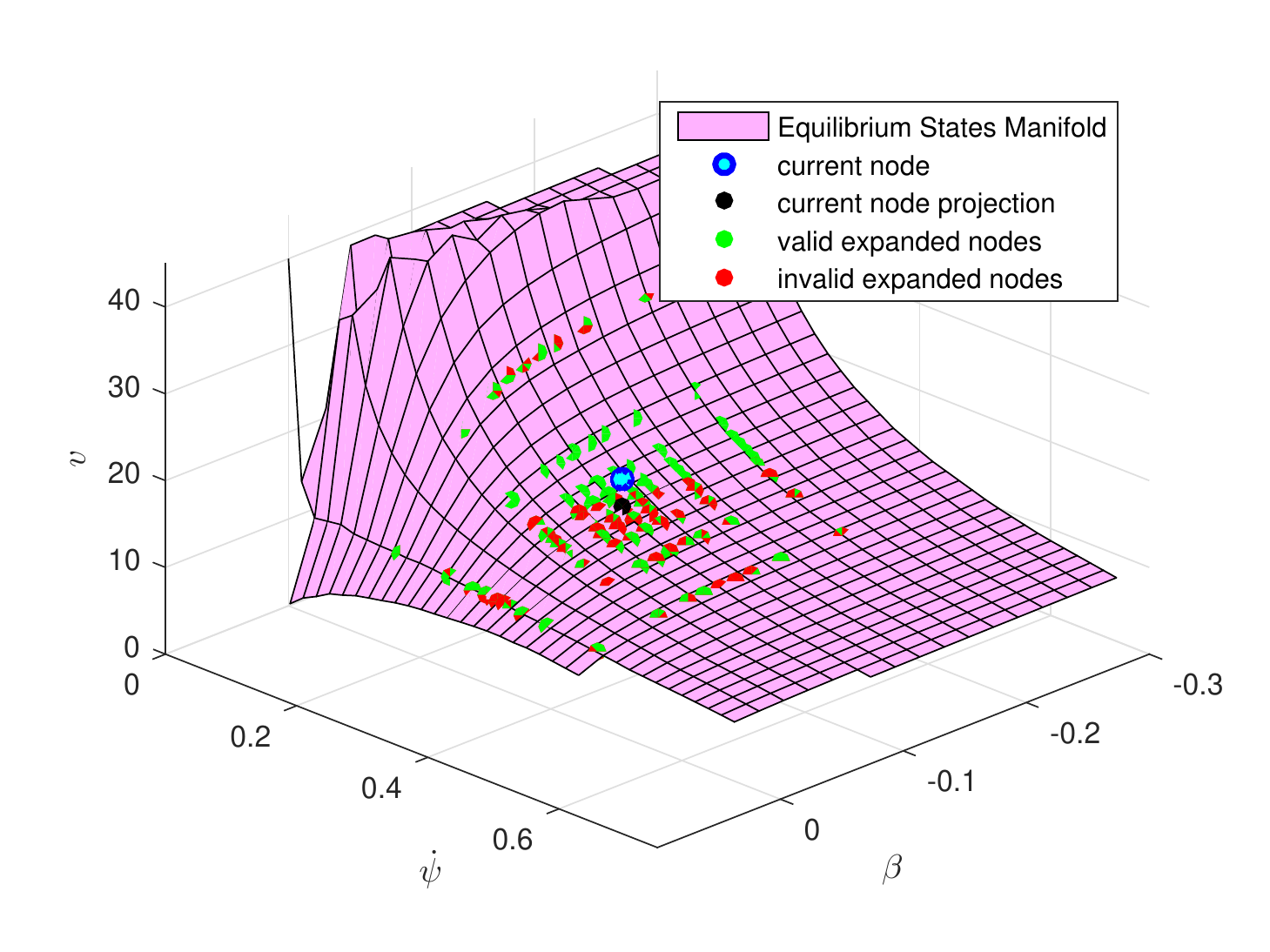}%
		\caption{ Expanding parent node $n$ to different child nodes $\childnodes$ by sampling on the ESM. }
		\label{fig:ESMsample}%
	\end{figure}

	During cornering, motion primitives are generated based on the Equilibrium State Manifold $\esm$, generated offline (as it is explained in Section \ref{subsec:v_manifold}), consisting of states $\stateESM = ( v, \beta, \dot\psi)$ each representing feasible solution for continuous drifting with different radii (\lq\lq donut drifting'').
	A race track is composed of different sections, with varying  curvature radii (as well as straight segments). Therefore, a continuous transition between different steady-states is desired.
	Based on the current node $\node$ (respective state $\CSTATE$), respective steady-state $\stateESM$ on ESM ($\esm$) is obtained by projecting onto the manifold $\esm$. From ESM ($\esm$) several reachable steady-states are sampled in the neighborhood of $\stateESM$ and kinematic model \eqref{eq:kinematic_model} is used to simulate the evolution of the additional states ($x, y, \psi$), assuming the linear transition between the current and sampled neighboring steady-states, effectively generating steady-state drifting motion primitives (with full state trajectories). 
	The final state of each motion primitive is used to generate one child node $\node'$.
	Neighboring steady-states are obtained by sampling the $\beta \times \dot\psi$ space around the current $\stateESM$ (with values $(\beta_0, \dot\psi_0)$), with the density of the samples decreasing as the distance from the $\stateESM$ increases (see Figure \ref{fig:ESMsample}). 
	This approach of sampling in $\esm$ effectively reduced the problem from sampling trajectories in 7-dimensional space to sampling the point in 2-dimensional space.
	For us, this pattern of sampling showed good results. However, different sampling patterns could be also employed. 
	The number of samples is a hyperparameter, which impacts considerably the performance of the search. As the number of samples is increased smoother trajectories can be planned, but the branching factor of the tree increases so computation time increases exponentially. Therefore a fine trade-off between computation time and sufficient space exploration is needed. 
	
	Equation \eqref{eq:derivatives_0}, used as a condition to generate the ESM, assumes that the rates of change of states are equal to zero. However, we need to transit between close states in order to generate trajectories with varying curvature in order to keep the vehicle on the road. This implies that this constraint (equation \eqref{eq:derivatives_0}) must be \lq\lq softened''.
	Still, it is important to keep it low, so limits on relative change must be set such as shown in \eqref{eq:Delta_assumption}. 
	\begin{subequations}
	\label{eq:Delta_assumption}
	\begin{align}
	\Delta v \leq \Delta v_\mathrm{max},\\
	\Delta\beta \leq \Delta \beta_\mathrm{max},\\
	\Delta\dot\psi \leq \dot\psi_\mathrm{max}.
	\end{align}
	\end{subequations}
	The smaller the deviations are, the closer the trajectory is to the ESM, therefore the model is more accurate. In practice, these limits are obtained experimentally by increasing them and detecting when trajectories become infeasible.
	
	In order to avoid generating and propagating an excessive amount of samples (to decrease the branching factor of the search), even before the nodes are checked for collision and removed, the following rules are considered:
	\begin{itemize}
	\item only equilibrium points defined within the surface in Figure \ref{fig:surface_interp} are considered. This also means  that the minimum reachable curvature radius is $R_\mathrm{c,min}=10m$;
	\item the (small) portion of the curve such that $\beta\cdot\dot{\psi}>0$ is neglected since equilibrium points in which $\beta$ and $\dot\psi$ have the same sign are associated with low-velocity conditions;
	\item a maximum velocity deviation $\Delta v$ between two successive nodes is defined, such that  $\frac{\Delta v}{Ts}<\accmax$, where $\accmax$ is the estimated maximum deceleration allowed on the given road surface.
	\end{itemize}
			

	\subsubsection{Close-to-straight driving mode} 
	
	When initial state $\CSTATE$ is close enough to the origin of the $\beta-\dot\psi$ plane, i.e., for $|\beta|<\beta_\mathrm{lin}$ and $|\dot\psi|<\dot\psi_\mathrm{lin}$, motion primitives are also generated according to a semi-linearized bicycle model, where the forces in \eqref{eq:nonlinear_model} are replaced with their linearized approximations $\linmodel$.	
	In order to generate different motion primitives, inputs are varied such that different values for steering wheel angle $\delta$ and the rear wheels slip $\lambda$ are equidistantly sampled within the ranges. Ranges are defined by $|\beta|<\beta_\mathrm{lin}$ and $|\dot\psi|<\dot\psi_\mathrm{lin}$ as in the equilibrium surfaces in Figure \ref{fig:surfaces_inputs}. In this way, multiple motion primitives are generated with different end velocities and turning radii.

	Finally, all expanded nodes are checked for a collision based on \eqref{eq:road_col} and \eqref{eq:Delta_assumption}, and 
	nodes that are in a collision are removed, leaving the remaining ones feasible regarding vehicle dynamics and driveability.

	\subsection{Heuristic function} 
	The heuristic function $h(n)$ is used to guide the search. It estimates the cost needed to travel from some node $n$ to the goal state (\textit{cost-to-go}). As it is shown by \citet{hart1968formal}, if the heuristic function is underestimating the exact cost-to-go, the A* search provides the optimal trajectory. For the shortest path search, the usual heuristic function is the Euclidean distance. On the other hand, to find the minimum lap time, the heuristic should estimate the distance that the vehicle can travel from the current node during the defined time horizon. It is optimistic to assume that the vehicle accelerates (with maximum acceleration) in the direction of the road's central line until it reaches the maximum velocity, and then maintains it for the rest of the time horizon. Based on this velocity trajectory, the maximum travel distance can be computed and used as a heuristic.
	
	In order to bias exploration towards the preferred motions and improve robustness, on the cost of sacrificing theoretical optimality, the heuristic function is augmented considering, among others:
	\begin{itemize}
	\item a \lq\lq dynamic states evolution'' cost, which helps limit the rate of change of the references $v$, $\beta$, $\dot\psi$, in order to obtain smooth trajectories and improve closed-loop state tracking;
	\item penalization for trajectories approaching the roadside;
	\item penalization of the nodes with fewer siblings, thus biasing the search to avoid regions where only a few trajectories are feasible.
	\end{itemize}

	\subsection{Illustrative example} 

\begin{figure}
	\includegraphics[trim=-70 0 -70 0,clip, width=\linewidth]{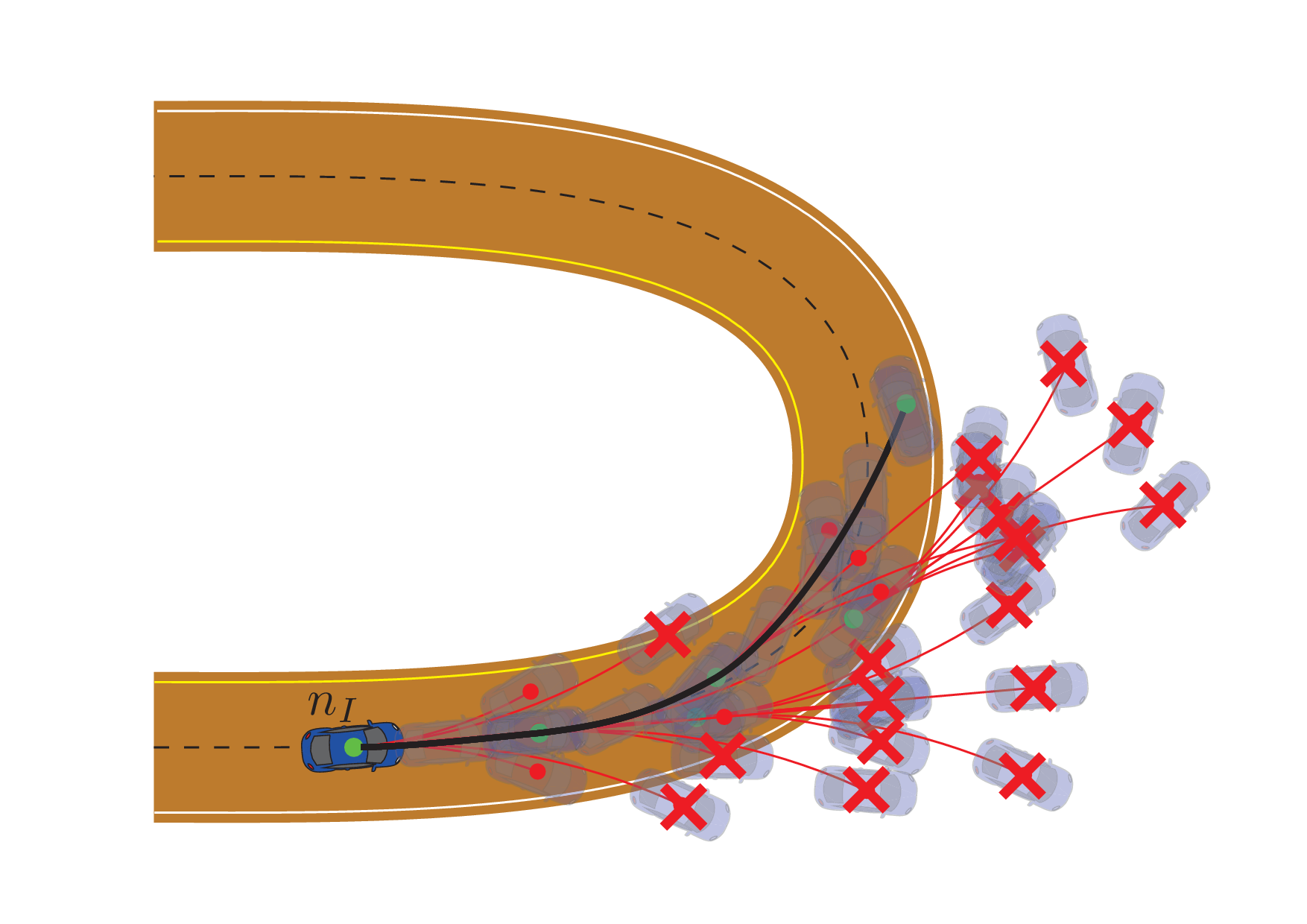}
	\setlength{\abovecaptionskip}{-7pt}
	\caption{Illustration of the principles of operation.}
	\label{fig:alg_illu}
	\end{figure}

	Constructed in this way, with presented components, SBMP can deal with nonlinear and hybrid vehicle models and plan for agile automated driving trajectories in a TAMP fashion. The method is generalizable to complex driving situations (arbitrary combinations of right and left curves and straight paths). Figure \ref{fig:alg_illu} illustrates one such example. The vehicle has to go into drifting mode to be able to drive through the sharp turn optimally. If drifting mode is not considered, the vehicle has to slow down significantly in order to stay on the road. As can be seen, many motion primitives lead the vehicle off the road and are therefore removed. Search is continued until some trajectory is found that keeps the vehicle on the road for the whole horizon.
	
	\section{Experimentation} 
	\label{sec:exper}

%
%
%

\begin{figure}
	\centering
	\subfigure{
		\frame{\includegraphics[trim={190pt 85pt 136pt 54pt},clip, width=1\columnwidth]{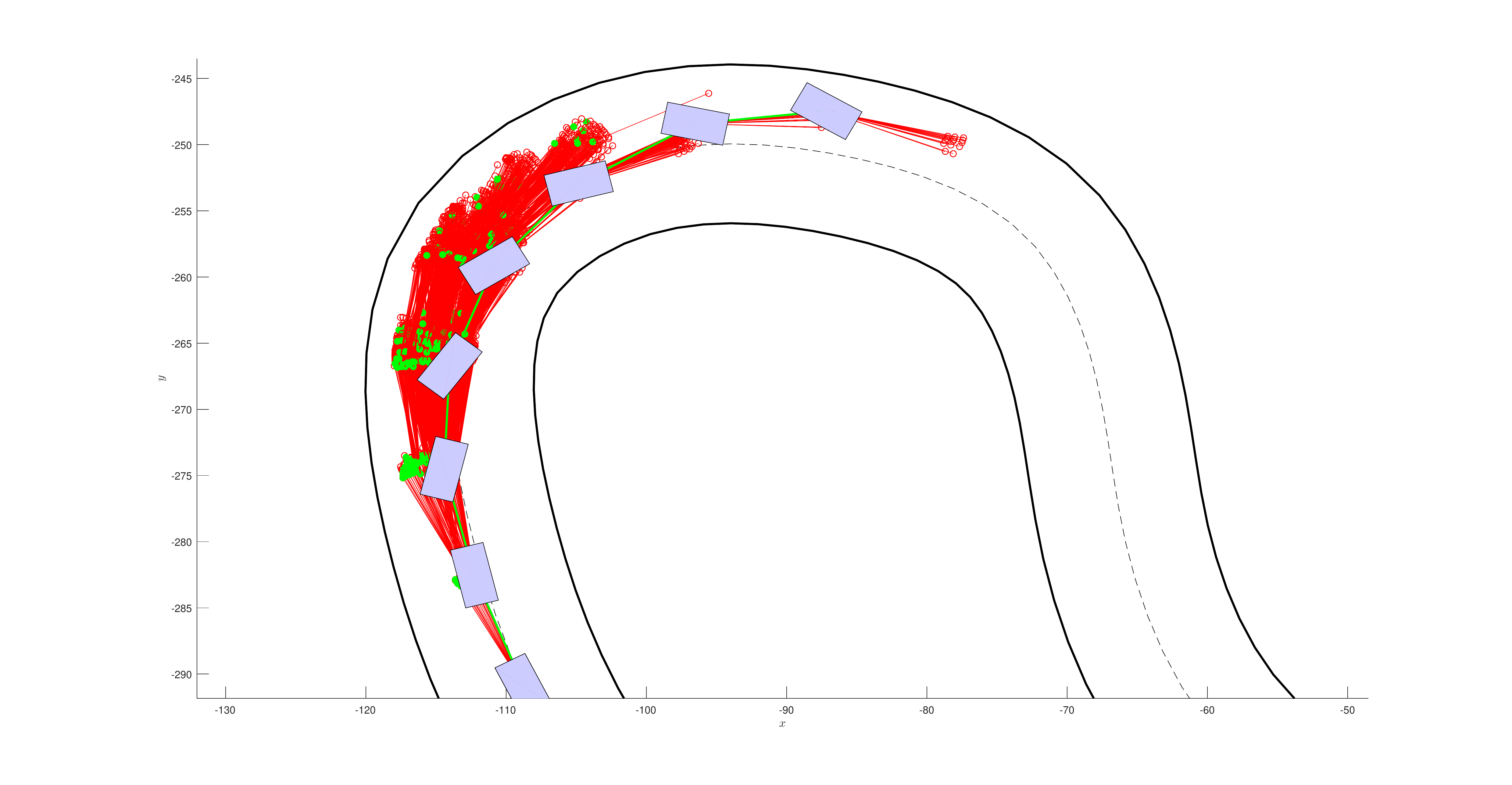}}%
	}\hfil
	\subfigure{
			\frame{\includegraphics[trim={190pt 85pt 136pt 54pt},clip,width=1\columnwidth]{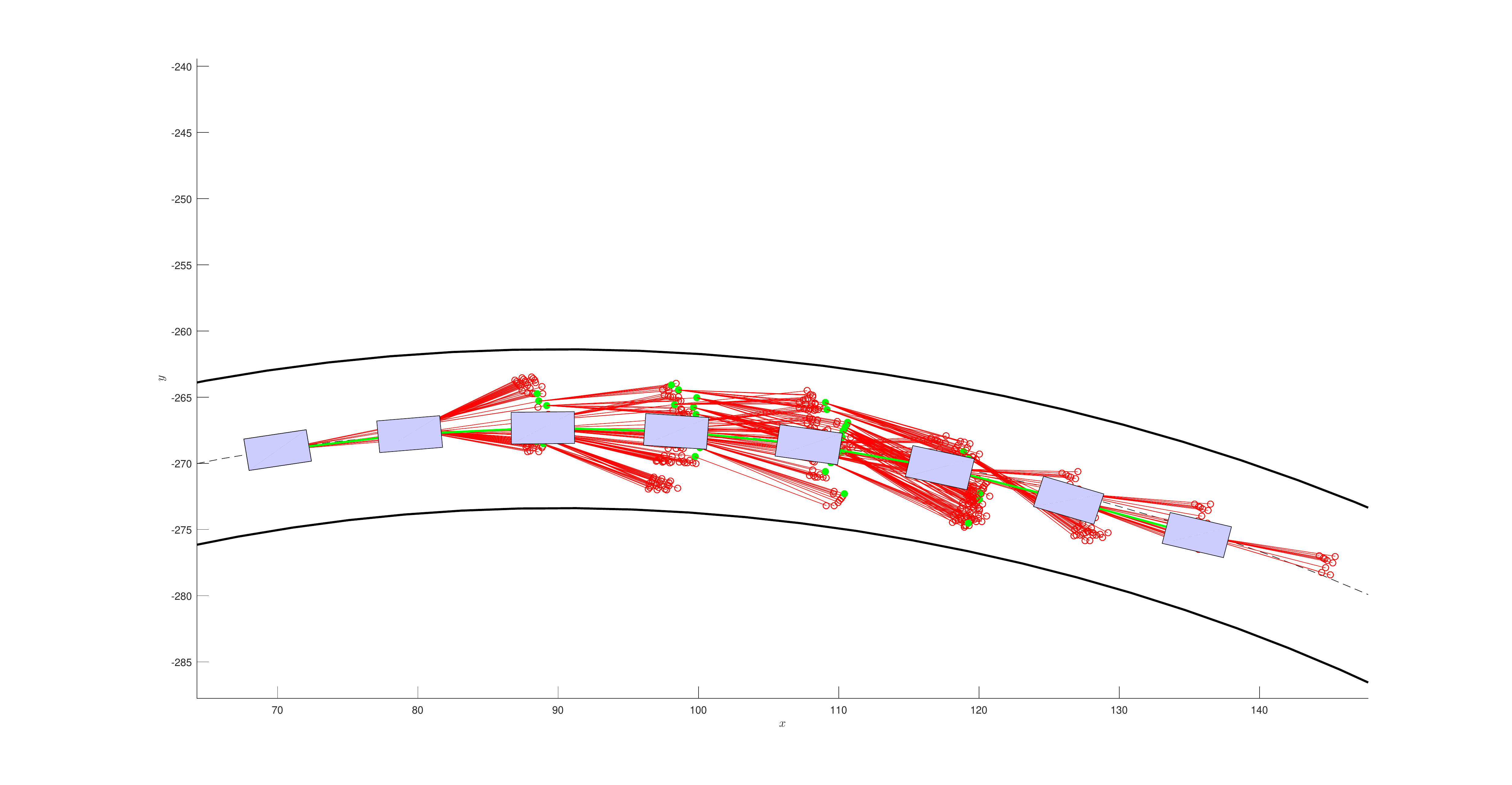}}
		}
		\caption{Graphical representation of the trajectories exploration in U-turn (top) and wide turn (bottom).}
		\label{fig:traj_exploration}%
	\end{figure}

\begin{figure}
		\centering
		\includegraphics[trim={2.5cm 0.5cm 2.5cm -0.5cm},clip,width=.92\columnwidth]{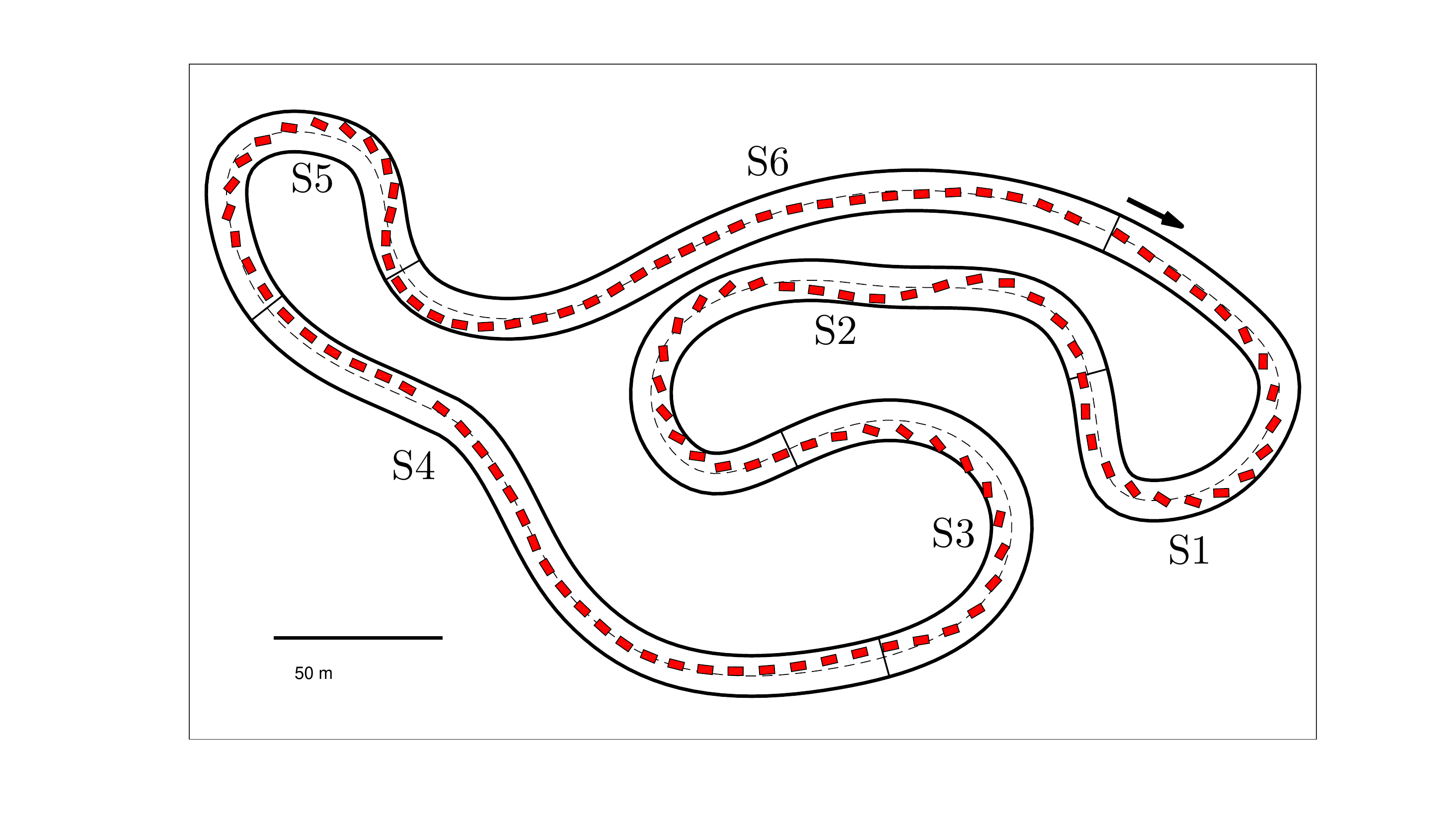}
		\caption{Obtained driving trajectory over the full test circuit.}
		\label{fig:trajectories}%
	\end{figure}

	The presented SBMP framework was adapted for the agile automated driving use case and implemented in the Matlab/SIMULINK environment. 
	As mentioned before, the established approach in motion planning for automated driving is to start re-planning from some future state from the previous plan as long as there is no large tracking deviation from the planned motion \citep{ziegler2014trajectory}. Therefore, we first verify planner performance assuming perfect actuation, i.e., the actual vehicle dynamical states/positions match the ones planned at the previous iteration. This is also important, as the focus of this work is on computationally efficient trajectory generation. Additionally, to prove that planned trajectories are feasible in the real system, we also show the performance of planned trajectory tracking in a closed loop using controllers on the full nonlinear vehicle model.
	For verification purposes, an artificially mixed circuit was used, characterized by slippery conditions (gravel), which contains several road sections of varying curvature radii, as can be seen in Figure \ref{fig:trajectories}. 	
	The proposed planner manages to find the appropriate vehicle trajectory for driving on the track. 
	An example of the algorithm exploration behavior is shown in Figure \ref{fig:traj_exploration} in the case of a U-turn and of a wider curve.
	The explored branches are represented by the red links, and the closed nodes are marked as green. The light-blue car frames represent the optimal vehicle states (see Figure \ref{fig:traj_exploration}).
	
\begin{figure*}
		\centering
		\begin{tabularx}{\linewidth}{|X|X|X|X|X|X|X|}
			\hline
			\includegraphics[trim={11cm 0 11cm 0},clip,width=.25\columnwidth]{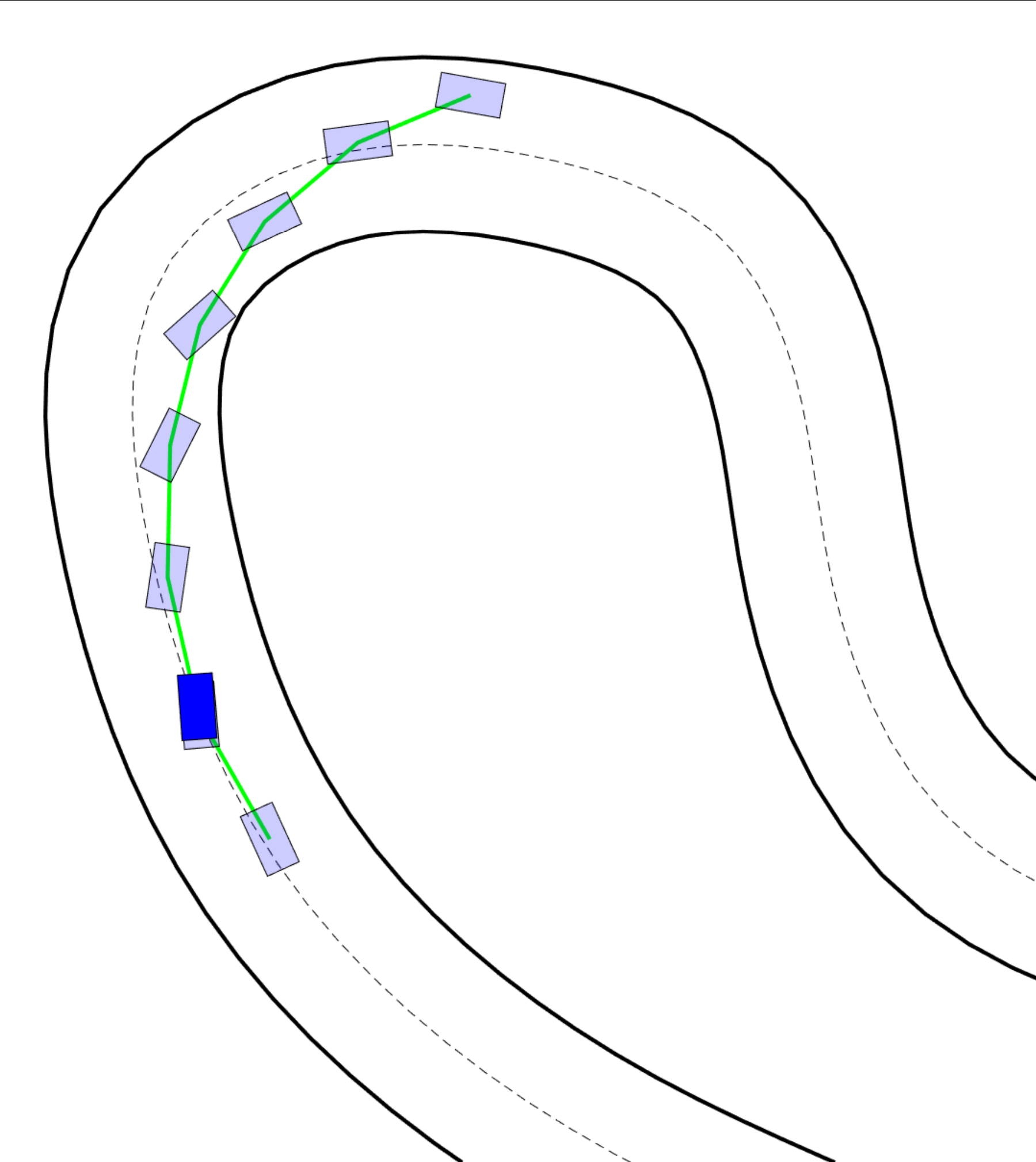} &
			\includegraphics[trim={11cm 0 11cm 0},clip,width=.25\columnwidth]{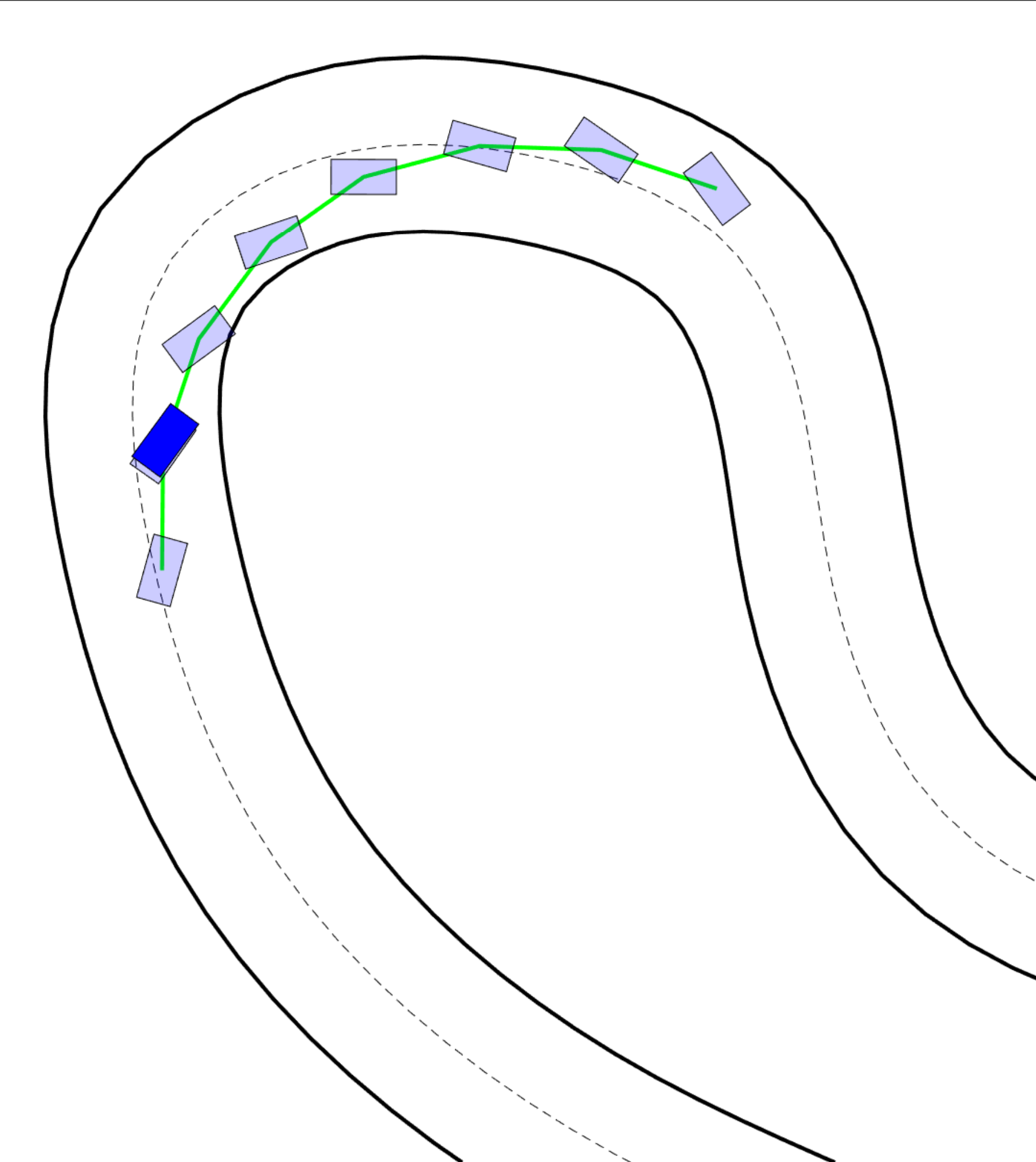} &
			\includegraphics[trim={11cm 0 11cm 0},clip,width=.25\columnwidth]{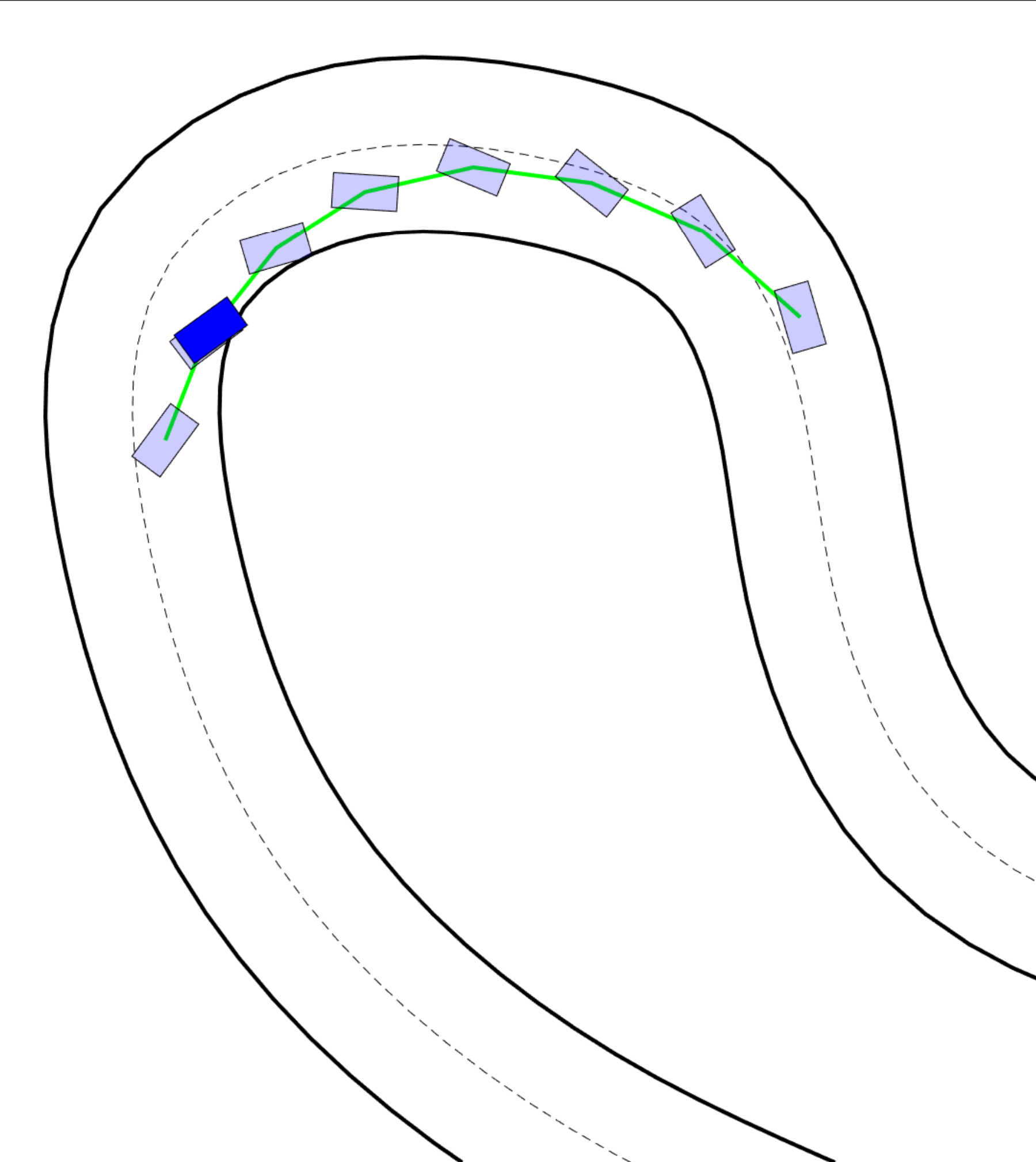} &
			\includegraphics[trim={11cm 0 11cm 0},clip,width=.25\columnwidth]{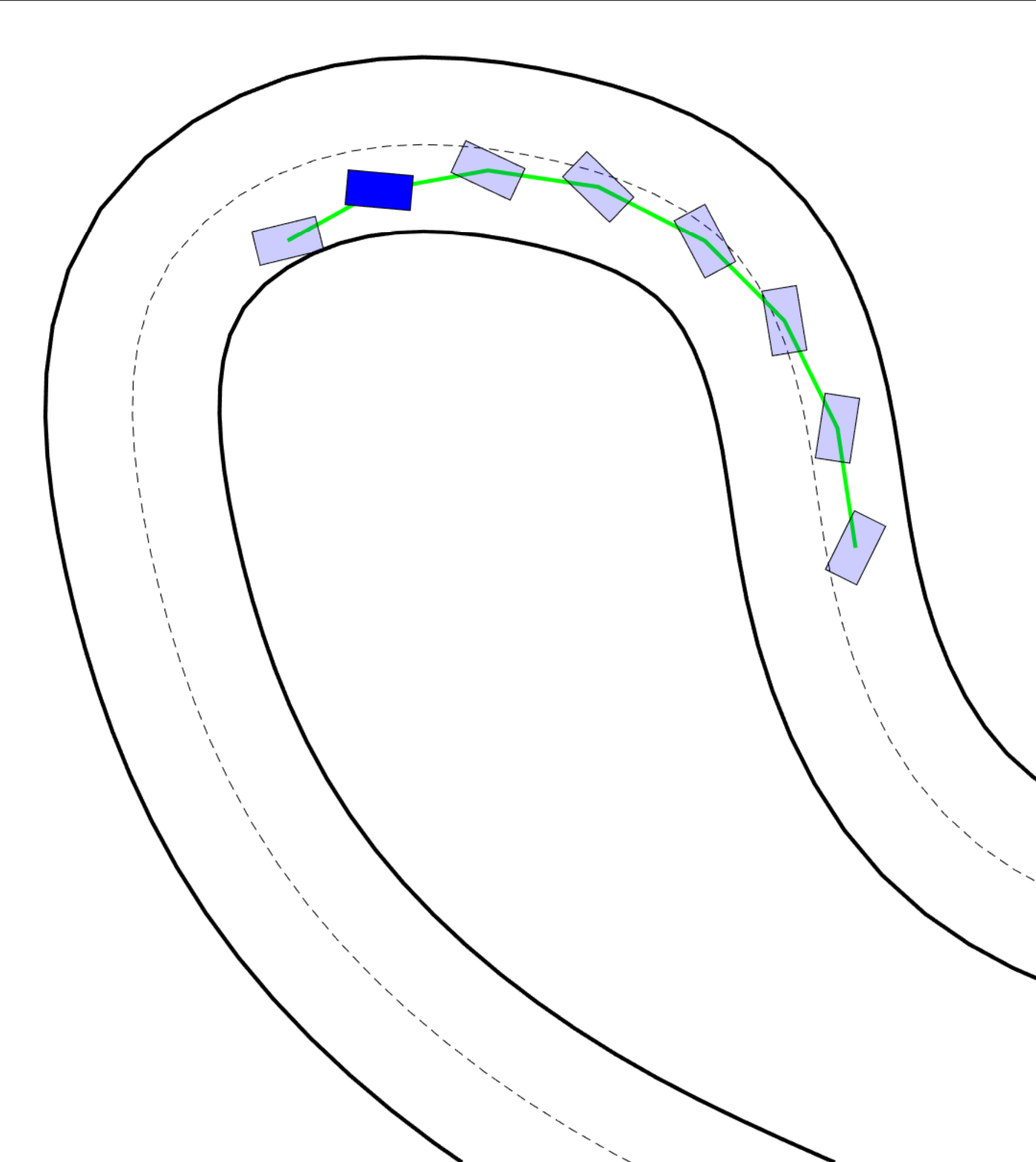} &
			\includegraphics[trim={11cm 0 11cm 0},clip,width=.25\columnwidth]{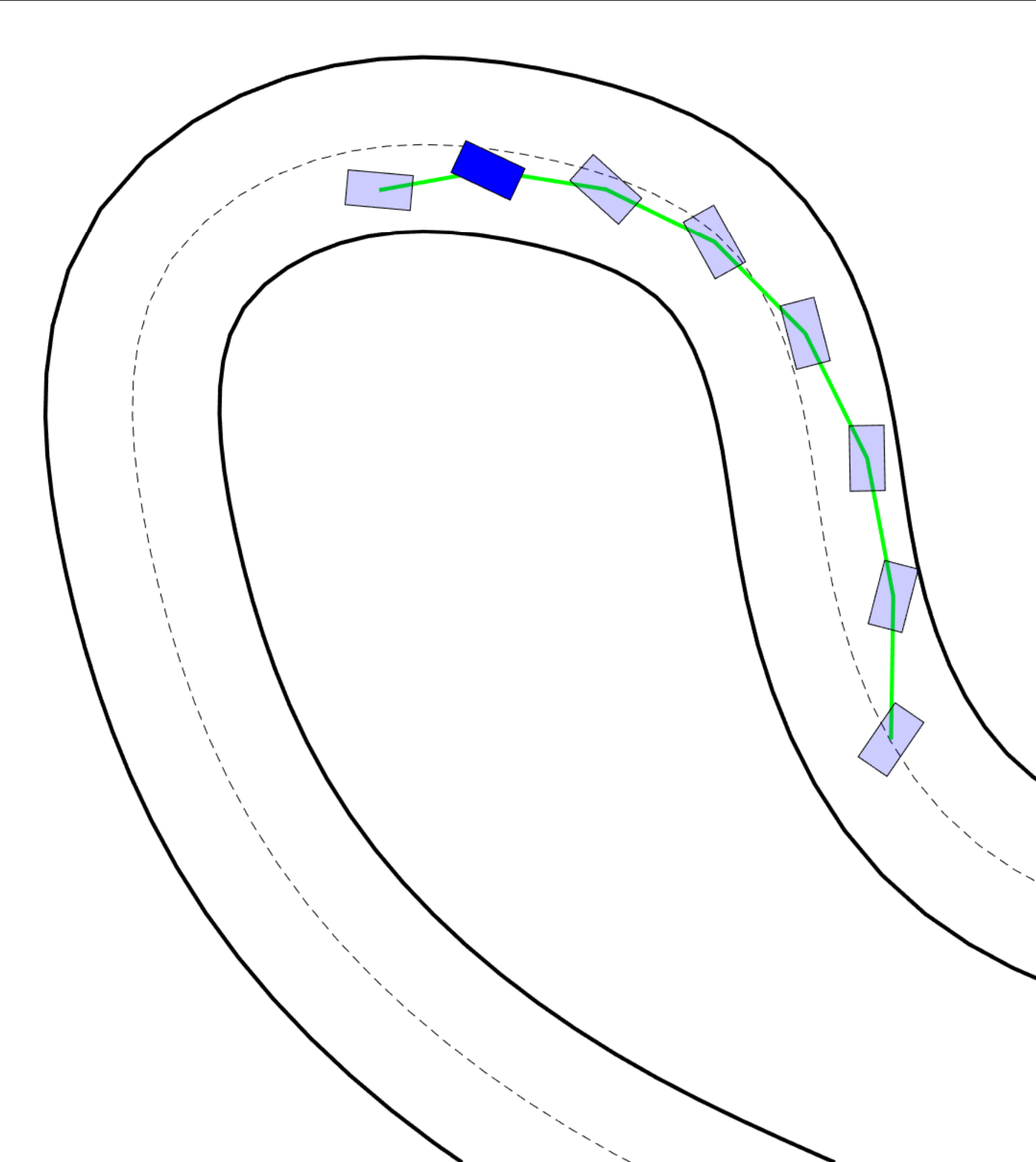} &
			\includegraphics[trim={11cm 0 11cm 0},clip,width=.25\columnwidth]{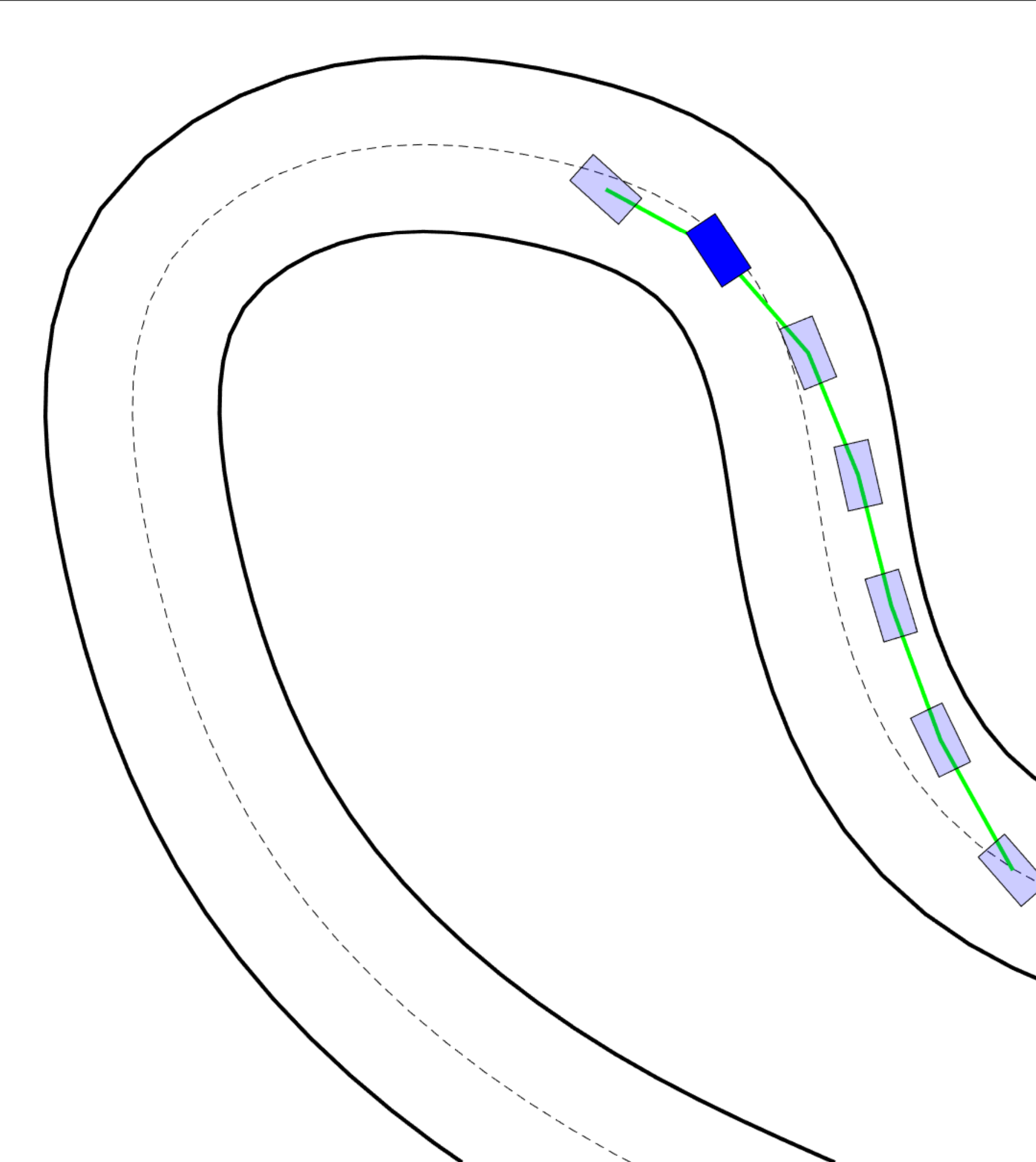} &
			\includegraphics[trim={11cm 0 11cm 0},clip,width=.25\columnwidth]{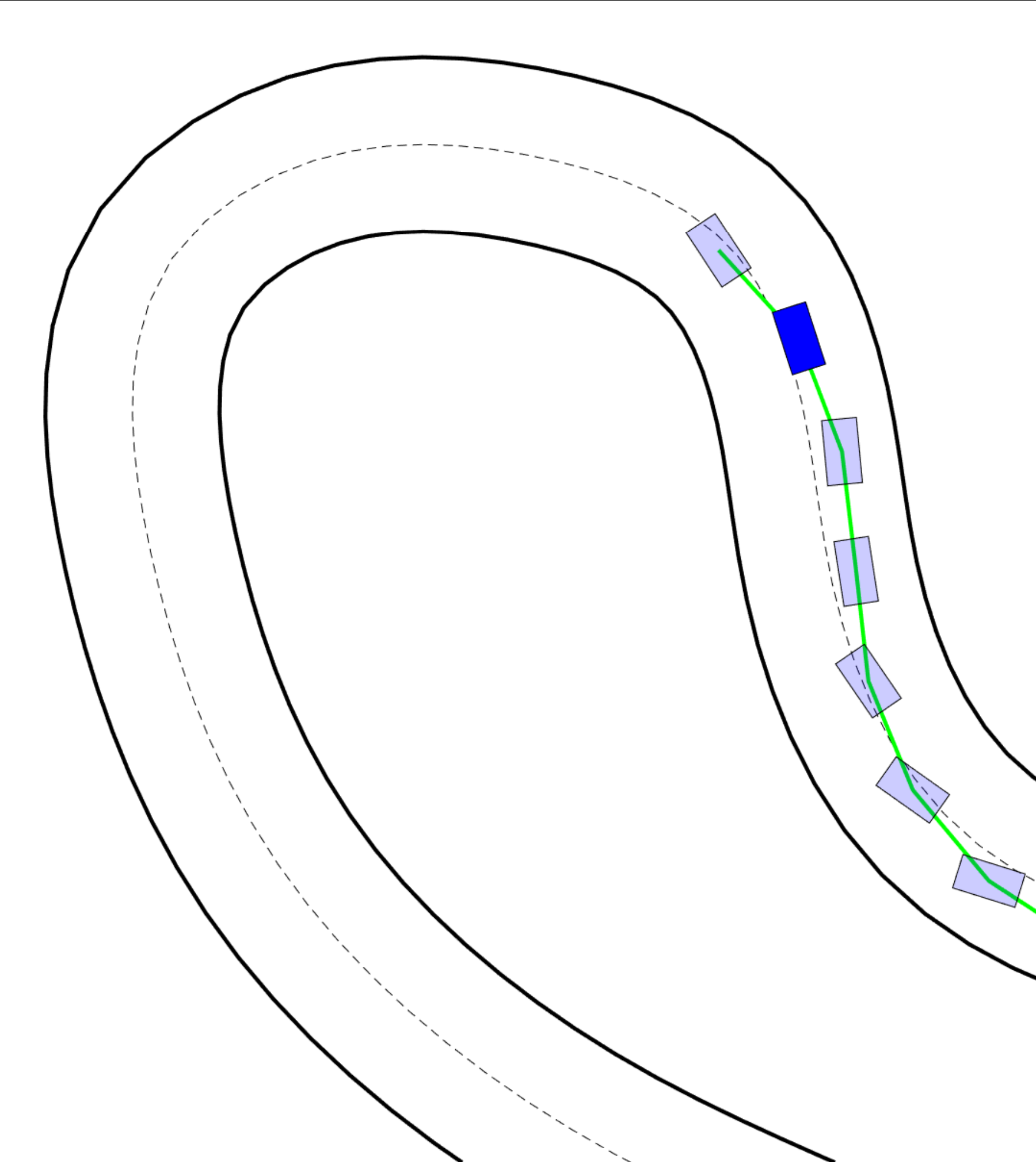} \\						
			\hline
		\end{tabularx}
		\caption{U-turn maneuver: consecutive frames.}
		\label{fig:frames}%
	\end{figure*}

	In Figure \ref{fig:frames}, several frames of the same maneuver are shown (the top left turn in the track illustrated in Figure \ref{fig:trajectories}). From these, it is possible to get an insight into how the optimal trajectory is re-planned, at each iteration, based on the current position. Given the nature of the receding horizon approach, it is not guaranteed (nor preferred) that all or part of the previously computed trajectory are kept in the next iteration. In fact, while in the first step the trajectory approaches ``dangerously'' the side of the road, in the next two steps the trajectory is incrementally improved, thanks to the fact that the exploration of such a portion of the track is now being evaluated in earlier nodes.

\begin{figure}
		\centering
		\includegraphics[trim={0.7cm 0.2cm 1.5cm 0.3cm},clip,width=.92\columnwidth]{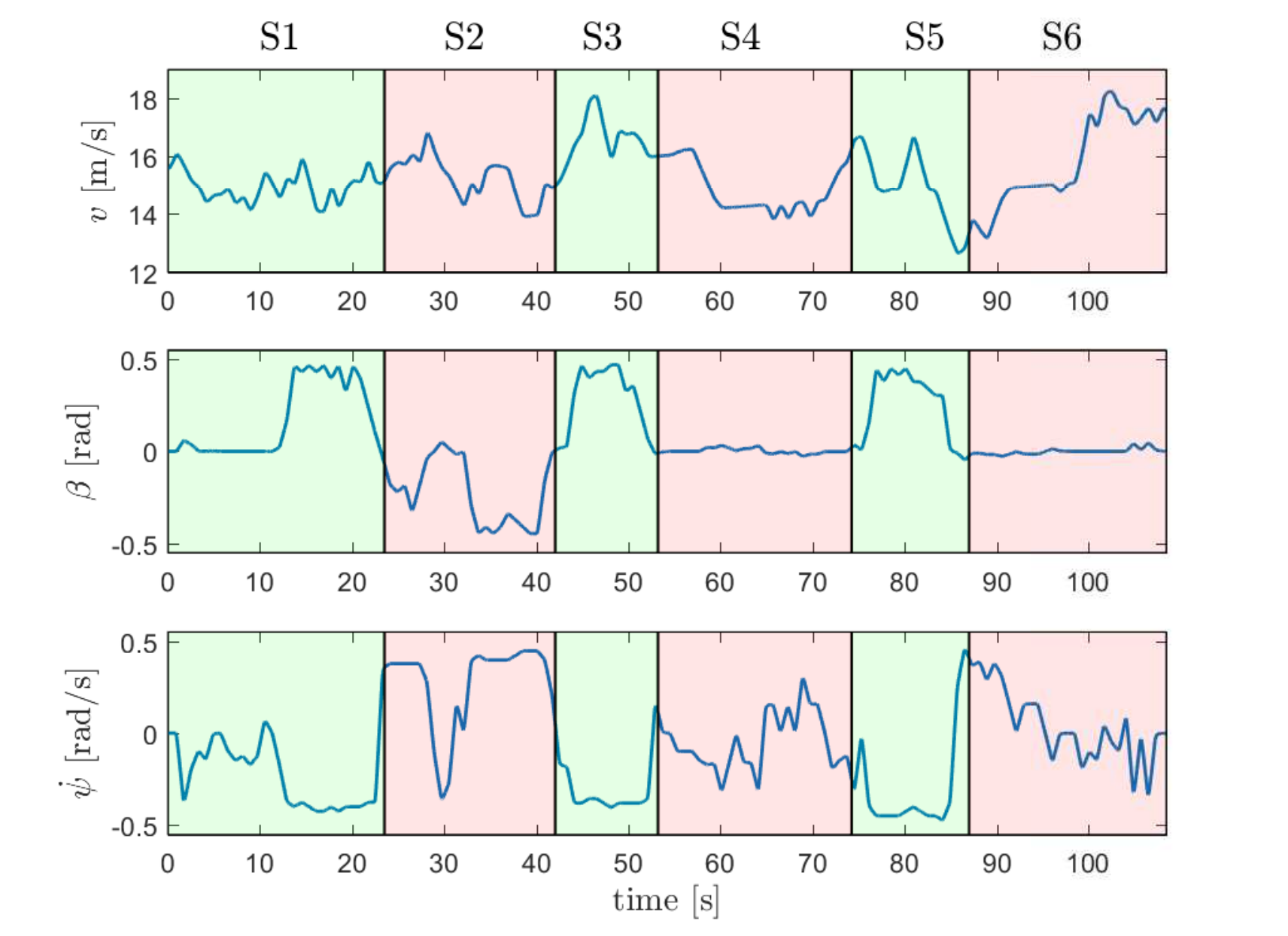}%
		\caption{Reference dynamical states over the full test circuit.}
		\label{fig:states_graph}%
	\end{figure}

	The dynamical states, which represent the output of the trajectory generation, are depicted in Figure \ref{fig:states_graph}. One can see how the generated references are varied smoothly, in particular in terms of $v$ and $\beta$, which are the quantities characterized by slower actuation dynamics.
	Moreover, it is possible to distinguish clearly 4 intervals in which the optimal generated maneuver is a 'drift' one with $\beta>0.4 \text{rad}$.
	These same intervals can be distinguished in Figure \ref{fig:trajectories}, where the overall trajectory on the considered $10 \text{m}$-wide track can be evaluated.

	For validating the advantage of using multiple modes (i.e., drifting) we benchmark our approach to other state-of-the-art approaches that do not use drifting mode but can still drive full circle autonomously. Due to the low friction coefficient of the dirt road, we found other approaches like \citet{liniger2015optimization} and \citet{li2022autonomous} difficult to adapt for these conditions and achieve the full circuit driving. After extensive unsuccessful trials, we used our planner with disabled drifting mode. We found that solution as well representative of these approaches as it uses the model with similar fidelity, but also has a defined domain of the applicability from mode-feasibility-map $\feasmap$. Mode-feasibility-map enables it to know the limits of the model and provide feasible trajectories also for low friction conditions.
	SBMP without drifting is achieving an average velocity of approx. 8 m/s, while the average velocity of SBMP with both modes is approx 15 m/s. This demonstrates the advantage of using drifting on low-friction surfaces and achieving minimum-time cornering.

	\subsection{Controllers}
	For the execution of the motion plans, in this work, we use two different controllers. One for each of the modes. For steady-state cornering mode, we use a drifting controller \citep{goh2019towards}. And for close-to-straight driving, we use the path-following controller \citep{lu2018performance}, that includes a Sliding Mode Controller for lateral motion and a simple PI longitudinal controller that minimizes the weighted sum of velocity and position-lag error. We switch between these two controllers based on the mode selected by the planner.
	
\begin{figure}
		\centering
		\includegraphics[trim={0.0cm 0.0cm 0.0cm 0.0cm},clip,width=.92\columnwidth]{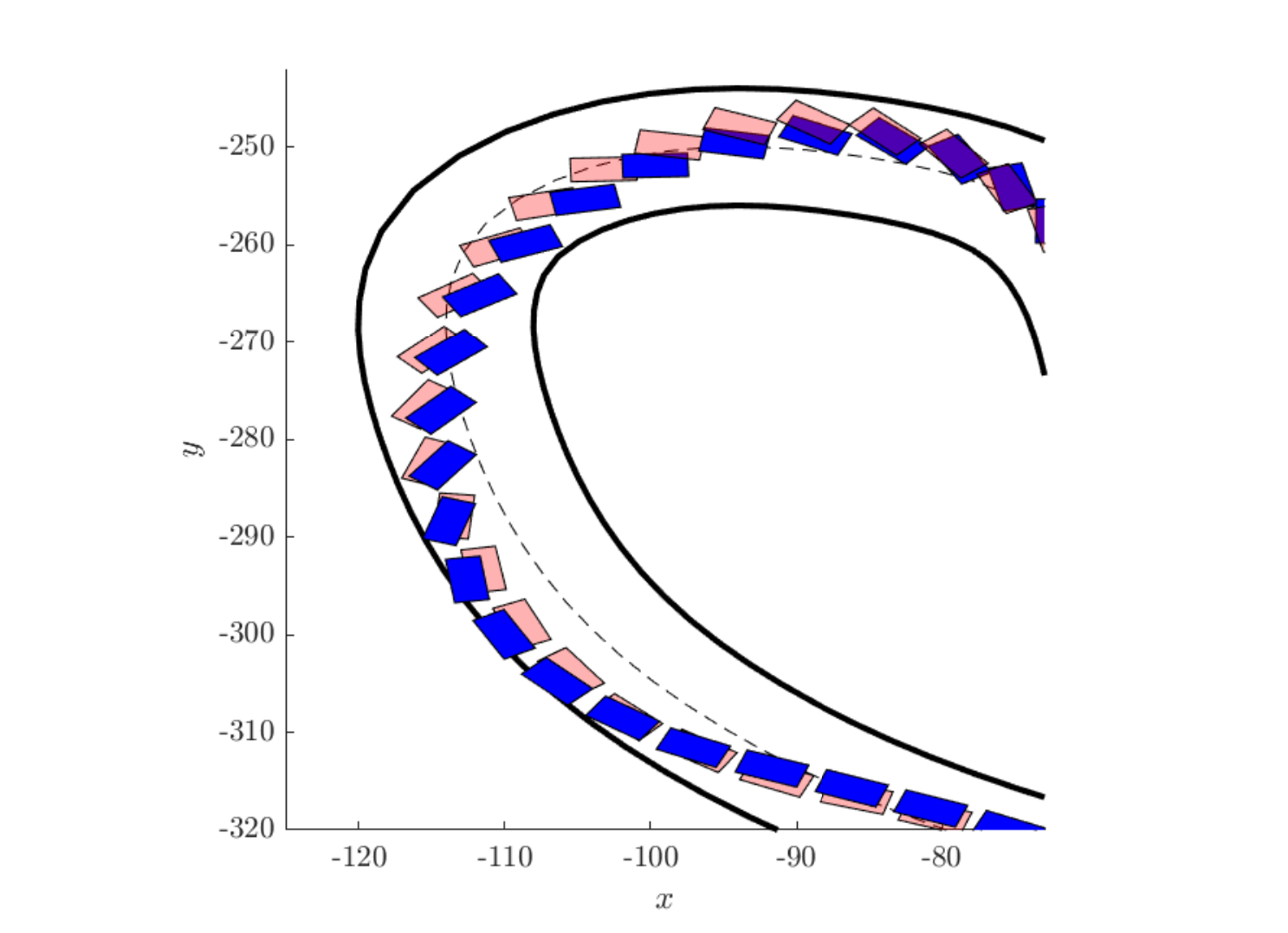}%
		\caption{Tracking reference trajectory using path tracking and drift controller.}
		\label{fig:tracking_driving_line}%
	\end{figure}

\begin{figure}
		\centering
		\includegraphics[trim={0.0cm 0.0cm 0.0cm 0.0cm},clip,width=.92\columnwidth]{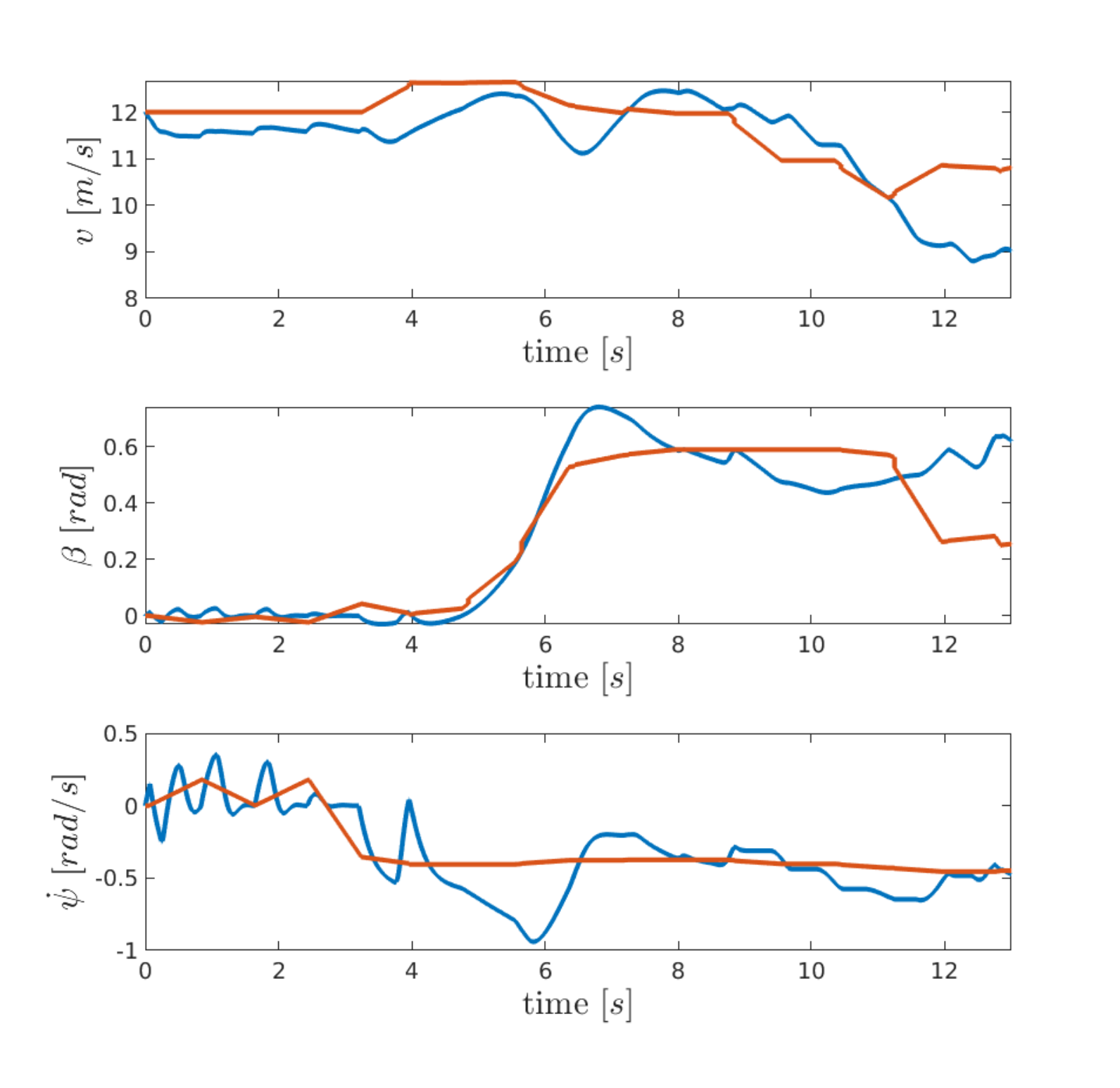}%
		\caption{Tracking reference trajectory using path tracking and drift controller (state trajectory).}
		\label{fig:tracking_states_graph}%
	\end{figure}

	As can be seen in Figure \ref{fig:tracking_driving_line}, controllers are robust enough and trajectories are feasible so the vehicle can drive through a very challenging curve. At around 4 s, mode is changed from close-to-straight driving to drifting. This is seen also on the side-slip angle $\beta$ in the Figure \ref{fig:tracking_states_graph}. The drifting controller successfully overtakes the control and continues through the curve. Although controllers are not tracking perfectly reference states, the final diving line is closely following the reference. 
	
	\subsection{Computational performance}
	
	It is well known that the computational complexity of the A* search depends on the quality of the heuristic function \citep{russell2021artificial}. In the worst case, when the heuristic function is not informative at all, the algorithm behaves as an exhaustive search, with exponential time complexity in the depth (in the order of $O(b^d)$). Where $b$ represents the branching factor and $d$ represents the depth.
    
	In our algorithm, the depth $d$ represents horizon length. More precisely, the horizon time of MPC divided by the time-step length of motion primitives.
	On the other hand, the branching factor $b$ represents the number of sampled motion primitives generated at each $\texttt{Expand}$ step (in the order of 100). 
    To achieve real-time algorithm performance, we utilized a hybrid A* approach that prunes generated motion primitives based on the discretized search space and practically reduces the branching factor. 
    An additional advantageous feature of our approach is that we design the algorithm as an anytime algorithm with a timeout. If some planning instance is harder to solve, by limiting the number of nodes, we  practically shorten the horizon so it can be solved faster. This provides a suboptimal solution, but the solution can be corrected again in the next re-planning step.
	
\begin{figure}
	\centering
	\subfigure{
		\frame{\includegraphics[width=1\columnwidth]{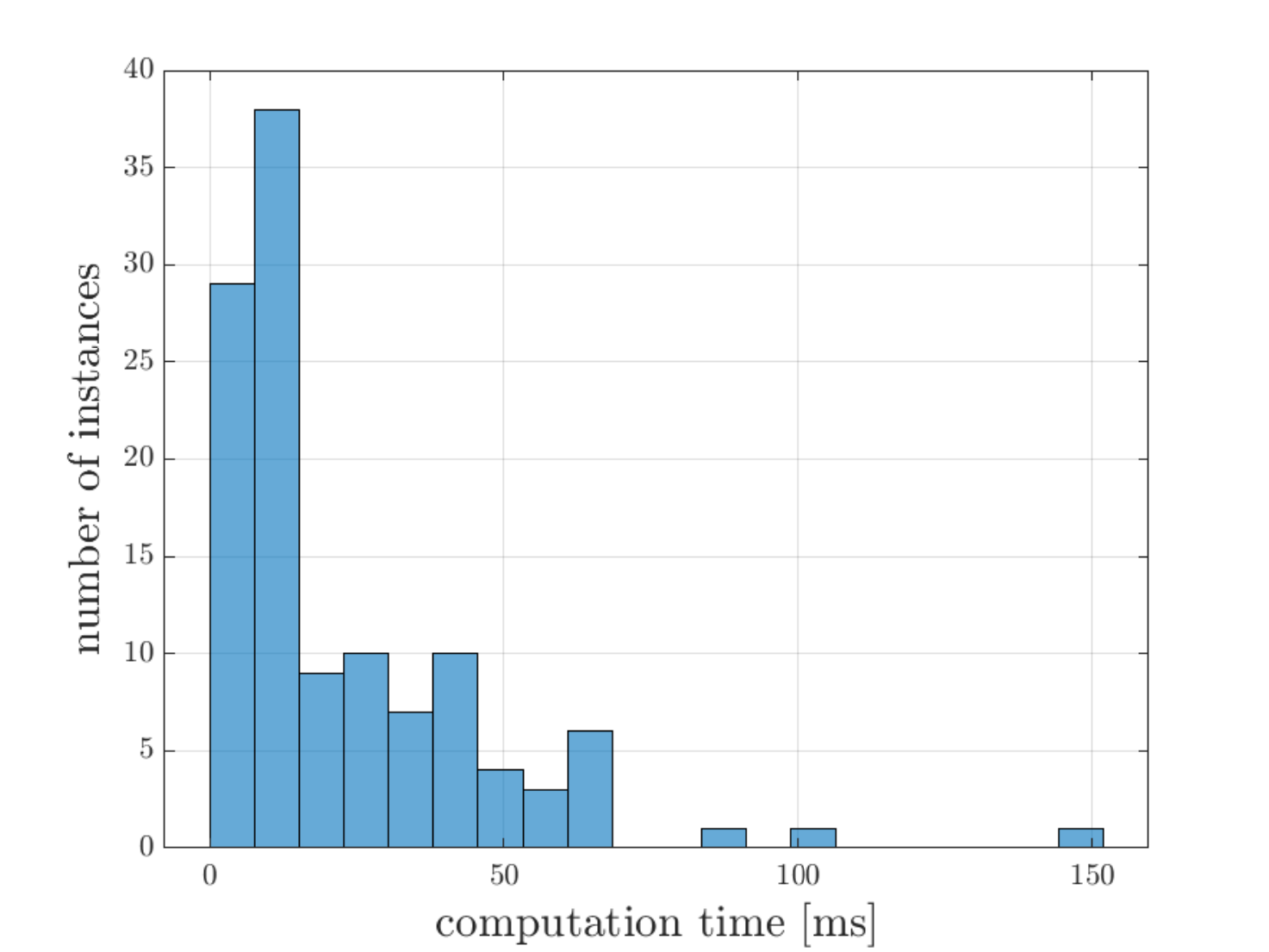}}%
	}\hfil
	\subfigure{
			\frame{\includegraphics[width=1\columnwidth]{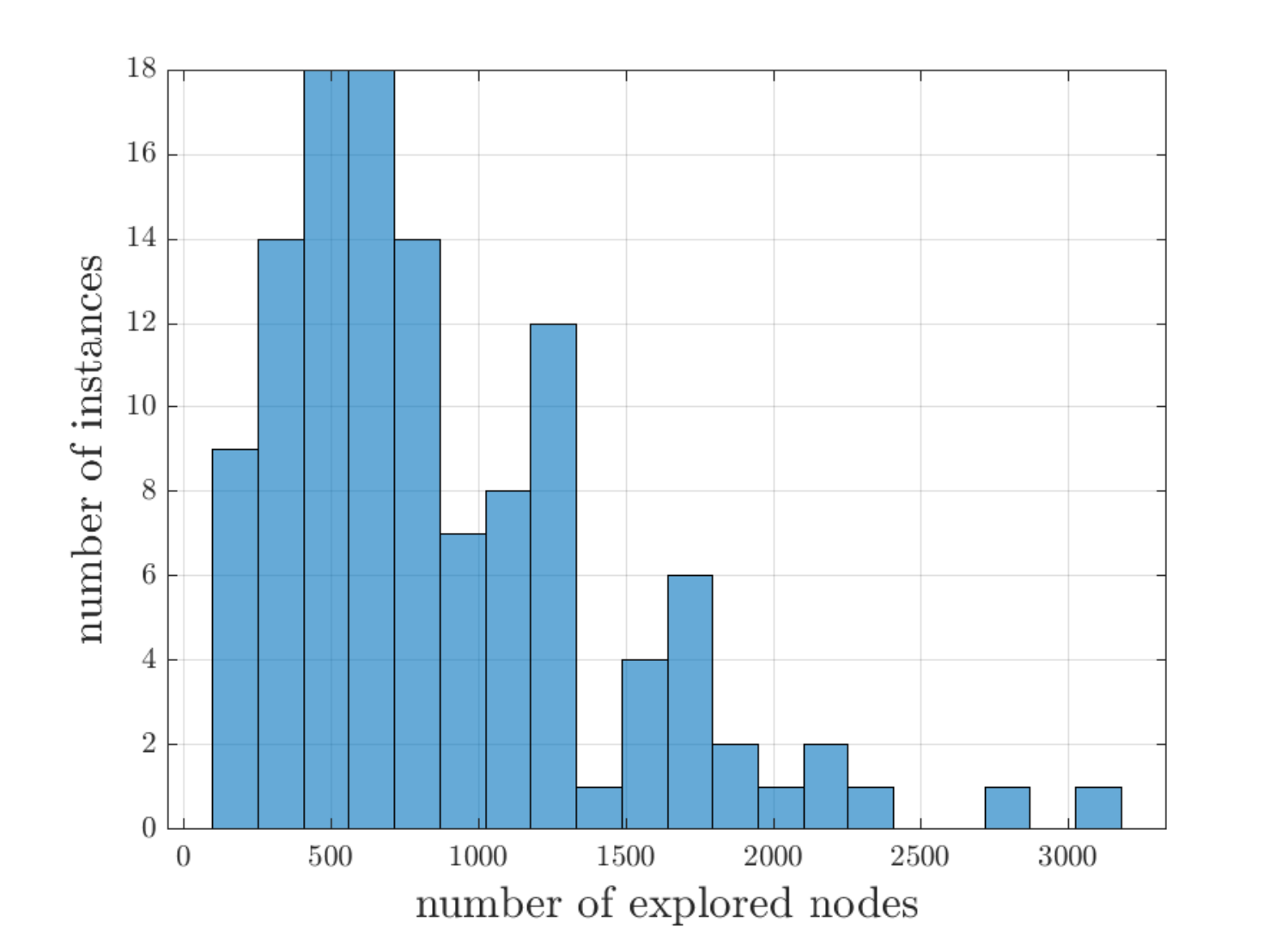}}
		}
		\caption{Histogram of computation times (top) and explored nodes (bottom) for the track.}
		\label{fig:historgram}%
	\end{figure}

	Besides theoretical computational complexity, practically, the usability of the algorithm very much depends on the constants in the complexity relation. Well-optimized implementation and appropriately tuned problem parameters can make it very practical. Besides the Matlab/SIMULINK implementation, previously mentioned, we validated an efficient C++ implementation to determine the ultimate practical computational performance of our approach. On the same simulated lap as before, the C++ SBMP planner was used in MPC fashion and computational performance for all SBMP planner calls is presented in Figure \ref{fig:historgram}. These are the results achieved on the computer with Intel i5 8th generation CPU, with 8 Gb RAM. As can be seen in the left figure, all planning computations are under $100 \text{ms}$ with a median time of $13.75 \text{m}s$ and mean time of $22.95 \text{ms}$. As can be seen in the right figure, all computations explored under $3500$ nodes to find the solution, with a median of $716$ nodes.
    
\begin{figure}
	\centering
	\subfigure{
		\frame{\includegraphics[width=1\columnwidth]{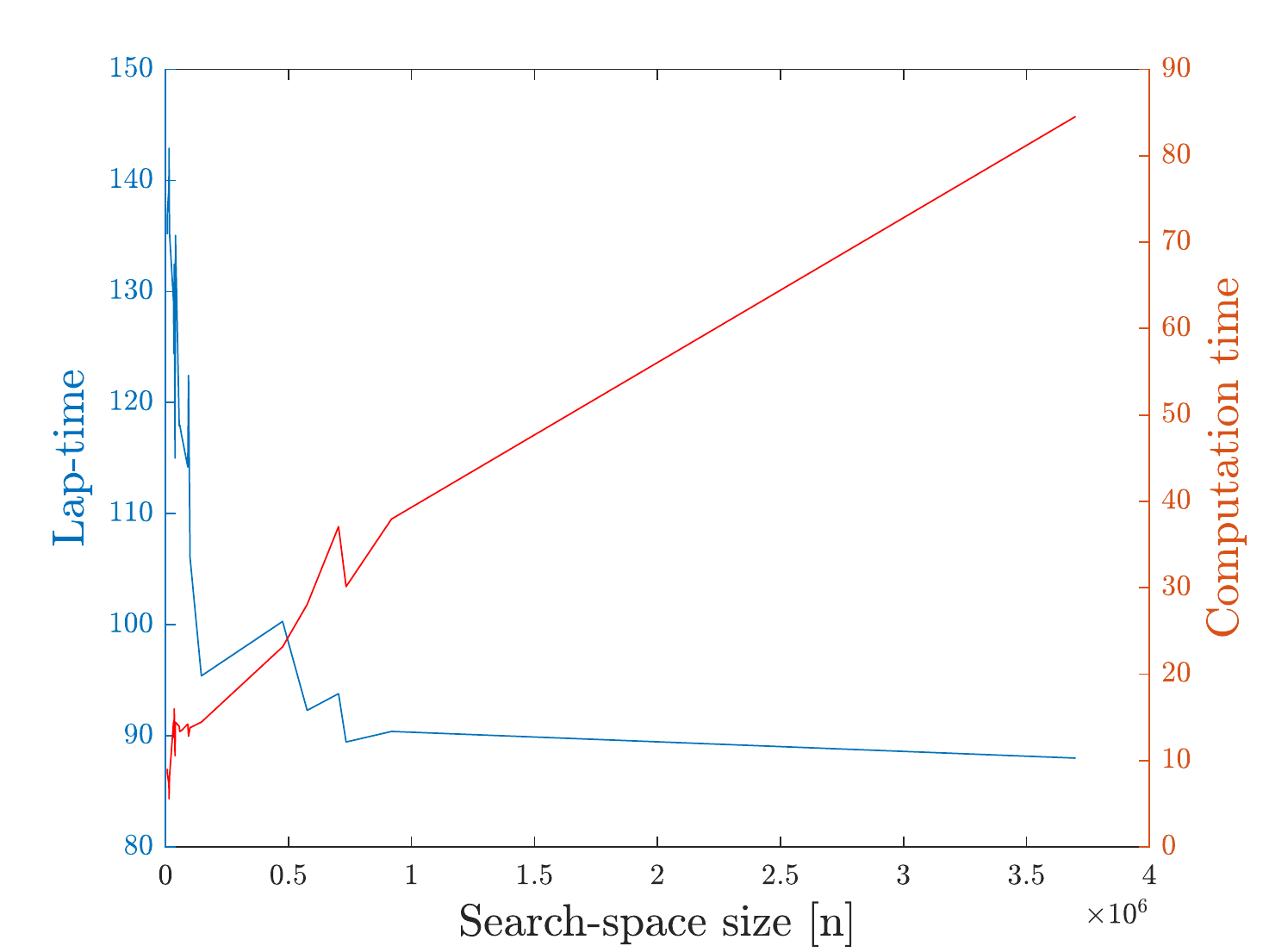}}%
	}\hfil
	\subfigure{
			\frame{\includegraphics[width=1\columnwidth]{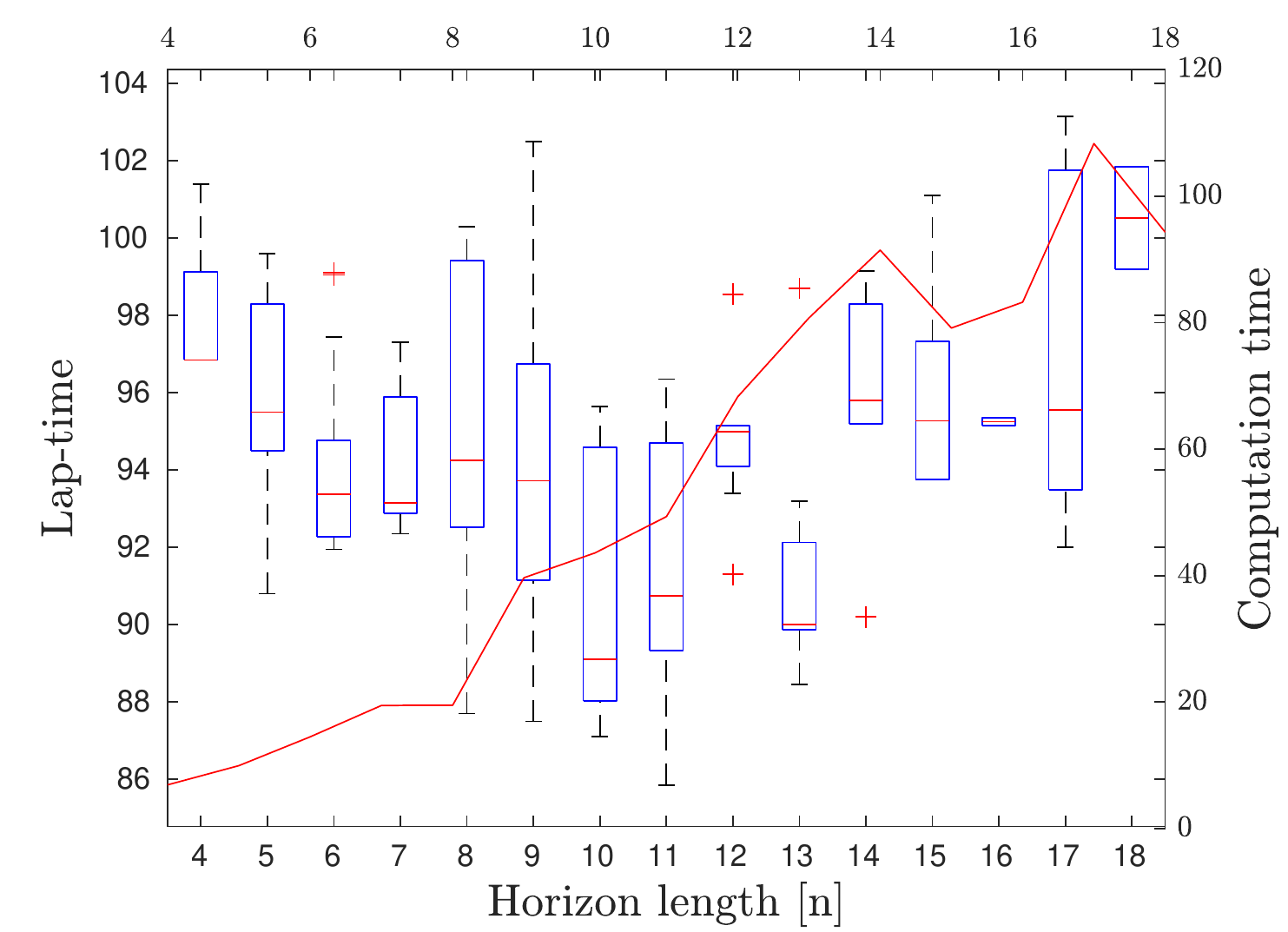}}
		}
		\caption{Computational time and lap time depending on the search-space size (top) and horizon length (bottom).}
		\label{fig:complexity}%
	\end{figure}

	Additionally, we performed an extensive simulation study to analyze the practical computational complexity of the algorithm in terms of horizon length and search space size. As mentioned earlier, there is a trade-off between computational time and solution quality measured in the lap time. Therefore we show both of them in Figure \ref{fig:complexity}.
	For analyzing the sensitivity on the horizon length, we preserve well-tuned search-space discretization and vary only the horizon length. To be able to keep the vehicle on the road (i.e. provide a sufficient planning horizon in the future), the time horizon in MPC is fixed. So, as we change the horizon length (number of motion primitives), we also adapt accordingly the time step length of motion primitives, so that their product is constant. 
	Experimental results from 139 laps are shown in Figure \ref{fig:complexity} (right). The results indicate that the computation time increases as we increase the horizon, following the exponential trend. There is no clear trend in the lap time and the mean time is rather constant with variations across multiple runs.
	It is worth noting that for horizons less than 4 steps, the vehicle is not able to drive the full lap without losing control.
 
	For analyzing the sensitivity on search-space size, we vary discretization steps for each of the state variables so we have different sizes of the grid that represent $\states$.
	Experimental results over 50 laps (each lap 50+ planning instances) are shown in Figure \ref{fig:complexity} (left). The results indicate that computational time increases as we increase the horizon, following the linear trend (on a selected range). On the other hand, the lap time marginally improves after cca 700k states. It is important to note that the results are slightly misleading as the search-space size actually scales exponentially with the number of discretization steps for state variables.
 
    This analysis provides a deeper insight into the computation complexity of our approach.
    It is important to highlight again that we show practically that close-to-optimal lap time can be achieved with acceptable computation times for a reasonable planner setting.
    %
	
	\section{Conclusions and Outlook} 
	\label{sec:concl}

	In this paper, we presented the SBMP, a novel A* search-based task and motion planning approach that enables agile automated driving on a slippery surface. The proposed method enables us to extend state-of-the-art approaches for drift-like driving from a steady-state drifting on a single curve to continuous driving on the arbitrary road, effectively entering (or exiting) drifting maneuvers and switching between right and left turns.
	The SBMP consists of tree search, efficient generation of dynamically feasible successor nodes based on motion primitives, and the mode-feasibility map that enables us to use multiple locally approximated models for motion primitives generation.
	In this way, SBMP treats this problem as TAMP and effectively decides \textit{when} to go into which \textit{mode} and \textit{how} to execute it.	
	The proposed method assumes that the vehicle parameters and the road surface properties are known to a certain degree, which allows to define a set of steady-state cornering maneuvers. 
	The method is evaluated on a mixed circuit characterized by slippery conditions (gravel), which contains several road sections of varying curvature radii $\curveradii$. 
	In several instances, due to the particular road surface considered, the optimal selected trajectory involves drifting, which in certain conditions ensures the maximum lateral acceleration.
	Such results demonstrate the capability of the proposed SBMP to generate feasible close-to-optimal trajectories on slippery conditions while considering a limited prediction horizon. Moreover, when considering U-turns with curvature radius as tight as 15m, trajectories are comparable in shape to the ones obtained by e.g., \citep{tavernini2013minimum}, when the full segment is optimized offline in order to find the minimum time optimal maneuver.

	Future research direction might utilize other ways to learn the Equilibrium State Manifold e.g., for real vehicles from human experts - Learning from Demonstration (LfD) or learning rapid generation of local car maneuvers similar to \citet{kicki2021learning}, and validated on a real vehicle similar to \citet{ajanovic2020validating}. 
	The sub-optimal policy could be further improved by learning from experience e.g., improving our base controller \citep{goh2019towards} with Residual Policy Learning \citep{silver2019residual}, a fully RL-based controller \citep{cai2020highspeed}, \citep{cai2020highspeed} or learning from corrections in an Interactive Imitation Learning fashion \citep{celemin2022interactive}. This might help to improve lap-time performance and to generalize to the distribution shift (e.g., changing tire-road conditions).
	Policy execution could be further improved by making a tighter connection between controllers and motion planning, e.g., by considering delays of controllers in the planning stage.
	Furthermore, the robustness might be improved by considering non-deterministic models.
	Future research directions might also consider more challenging scenarios such as multiple vehicles on the road in a race, where besides dynamics game-theoretic aspect should be considered in a minimax fashion. Our approach is well suited for non-deterministic and game theoretic extensions as it relies on the tree search. 
	Considering them might increase the computational complexity of the problem, but more advanced search algorithms or learning of heuristic functions similar to \citet{ajanovic2019novel} might help with that.
	Finally, as this approach is general, it would be useful to see its applicability to other agile robotic problems with similar structures and the extension of this approach with symbolic variables.

	
\section*{Acknowledgment}
The project leading to this study has received funding from the European Union's Horizon 2020 research and innovation programme under the Marie Sk\l{}odowska-Curie grant agreement No 675999, ITEAM project.
VIRTUAL VEHICLE Research Center is funded within the COMET - Competence Centers for Excellent Technologies - programme by the Austrian Federal Ministry for Transport, Innovation and Technology (BMVIT), the Federal Ministry of Science, Research and Economy (BMWFW), the Austrian Research Promotion Agency (FFG), the province of Styria and the Styrian Business Promotion Agency (SFG). The COMET programme is administrated by FFG. 
This research was partially supported by TAILOR, a project funded by EU Horizon 2020 research and innovation programme under GA No 952215.
We would like to thank Reviewers for taking the time and effort necessary to review the manuscript. We sincerely appreciate all valuable comments and suggestions, which helped us to improve the quality of the manuscript. Also, we would like to thank Rodrigo P\'{e}erez-Dattari and Pablo Borja for their comments on this work.

	
	
	\bibliographystyle{elsarticle-harv} 
	\bibliography{biblio.bib}

	\end{document}